\documentclass[letterpaper, 10pt, conference]{ieeeconf}  

\IEEEoverridecommandlockouts                             

\usepackage{amsmath} 
\usepackage{amssymb}  
\usepackage{array,makecell}
\usepackage{multirow}
\usepackage{multicol}
\usepackage{lipsum}
\usepackage{url}
\usepackage{booktabs} 
\usepackage{pifont}
\usepackage{pifont}
\newcommand{\cmark}{\ding{51}}%
\newcommand{\xmark}{\ding{55}}%
\usepackage{graphicx}
\usepackage{duckuments} 
\usepackage{adjustbox} 
\usepackage{scalerel} 
\usepackage{tikz} 
\usetikzlibrary{svg.path} 
\definecolor{orcidlogocol}{HTML}{A6CE39}
\tikzset{
  orcidlogo/.pic={
    \fill[orcidlogocol] svg{M256,128c0,70.7-57.3,128-128,128C57.3,256,0,198.7,0,128C0,57.3,57.3,0,128,0C198.7,0,256,57.3,256,128z};
    \fill[white] svg{M86.3,186.2H70.9V79.1h15.4v48.4V186.2z}
                 svg{M108.9,79.1h41.6c39.6,0,57,28.3,57,53.6c0,27.5-21.5,53.6-56.8,53.6h-41.8V79.1z M124.3,172.4h24.5c34.9,0,42.9-26.5,42.9-39.7c0-21.5-13.7-39.7-43.7-39.7h-23.7V172.4z}
                 svg{M88.7,56.8c0,5.5-4.5,10.1-10.1,10.1c-5.6,0-10.1-4.6-10.1-10.1c0-5.6,4.5-10.1,10.1-10.1C84.2,46.7,88.7,51.3,88.7,56.8z};
  }
}

\newcommand\orcidicon[1]{\href{https://orcid.org/#1}{\mbox{\scalerel*{
\begin{tikzpicture}[yscale=-1,transform shape]
\pic{orcidlogo};
\end{tikzpicture}
}{|}}}}

\usepackage[acronym]{glossaries}
\usepackage{acronyms}

\usepackage[hidelinks]{hyperref}
\usepackage[capitalize]{cleveref}
\crefname{figure}{Fig.}{figures}
\Crefname{figure}{Figure}{Figures}
\crefname{table}{Tbl.}{tables}
\Crefname{table}{Table}{tables}
\crefname{section}{Sect.}{sections}
\Crefname{section}{Section}{Sections}

\usepackage{textcomp}
\newcommand\copyrighttext{%
  \footnotesize \textcopyright 2024 IEEE. Personal use of this material is permitted.
  Permission from IEEE must be obtained for all other uses, in any current or future
  media, including reprinting/republishing this material for advertising or promotional
  purposes, creating new collective works, for resale or redistribution to servers or
  lists, or reuse of any copyrighted component of this work in other works.}
\newcommand\copyrightnotice{%
\begin{tikzpicture}[remember picture,overlay]
\node[anchor=south,yshift=10pt] at (current page.south) {\fbox{\parbox{\dimexpr\textwidth-\fboxsep-\fboxrule\relax}{\copyrighttext}}};
\end{tikzpicture}%
}

\title{\LARGE \bf
SubPipe: A Submarine Pipeline Inspection Dataset for Segmentation and Visual-inertial Localization
}

\begin{document}
\author{Olaya~Álvarez-Tuñón \orcidicon{0000-0003-3581-9481}, Luiza Ribeiro Marnet \orcidicon{0000-0001-6717-9306}, László Antal \orcidicon{0009-0005-4977-0959}, Martin Aubard \orcidicon{0009-0000-3070-8067}, \\ Maria Costa and Yury Brodskiy \orcidicon{0009-0002-0445-8126}
\thanks{O. Álvarez is with Artificial Intelligence in Robotics Laboratory (AiRLab), the Department of Electrical and Computer Engineering, Aarhus University, 8000 Aarhus C, Denmark (e-mail: olaya@ece.au.dk).
L. Ribeiro and Y. Brodskiy are with EIVA a/s, 8660 Skanderborg, Denmark (e-mails: \{lrm,ybr\}@eiva.com), L. Antal is with RWTH Aachen University, 52074 Aachen, Germany (e-mail: antal@informatik.rwth-aachen.de), M. Aubard and M. Costa are with OceanScan Marine Systems $\&$ Technology, 4450-718 Matosinhos, Portugal (e-mails: \{maubard,mariacosta\}@oceanscan-mst.com).
}%
}

\maketitle
\copyrightnotice
\thispagestyle{empty}
\pagestyle{empty}

\begin{abstract}
This paper presents SubPipe, an underwater dataset for SLAM, object detection, and image segmentation. 
SubPipe has been recorded using a \gls{LAUV}, operated by OceanScan MST, and carrying a sensor suite including two cameras, a side-scan sonar, and an inertial navigation system, among other sensors. The AUV has been deployed in a pipeline inspection environment with a submarine pipe partially covered by sand. The AUV's pose ground truth is estimated from the navigation sensors. The side-scan sonar and RGB images include object detection and segmentation annotations, respectively. State-of-the-art segmentation, object detection, and SLAM methods are benchmarked on SubPipe to demonstrate the dataset's challenges and opportunities for leveraging computer vision algorithms.
To the authors' knowledge, this is the first annotated underwater dataset providing a real pipeline inspection scenario. The dataset and experiments are publicly available online at \url{https://github.com/remaro-network/SubPipe-dataset}

\end{abstract}

\section{Introduction}
\label{sec:introduction}

Under the challenging imaging conditions that hinder the performance of computer vision algorithms, underwater vehicles' autonomy has been usually limited to sonar-based methods for localization and detection~\cite{rw:sonarwalldet}.
Nevertheless, the deep learning paradigm pushes the boundaries of computer vision-based algorithms underwater. Thus, the availability of data becomes the main limiting factor.

While other domains, such as autonomous driving, have long had a wide variety of datasets at their disposal \cite{cordts2016cityscapes, brostow2009semantic}, the availability in the underwater domain is limited by the difficulty of robot deployment and ground truth gathering. 
Offshore structures (such as pipelines) are suitable for \gls{AUV} deployment, as they require regular inspections. However, it is often unfeasible to open such data for public use because of privacy and security reasons.
Hence, most of the datasets available are recorded in the context of marine life monitoring \cite{rw:dataset:nautec, rw:dataset:suim} or archaeological inspection \cite{rw:dataset:aqualoc}. Moreover, these datasets are often divided between localization, segmentation, or object detection, and rarely provide ground truth for both. 

\begin{figure}[!t]
    \centering
    \includegraphics[width=\linewidth]{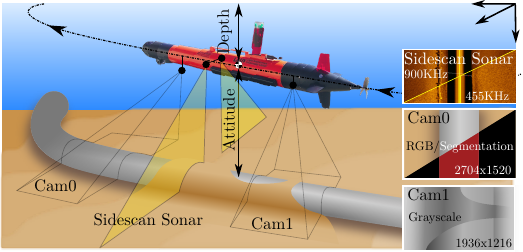}
    \caption{SubPipe has been recorded during a pipeline inspection mission with OceanScan's LAUV. The recorded data includes two monocular cameras (one monochrome camera and one RGB) with pipe segmentation annotations for the latter one; side-scan sonar images with bounding box annotations of the pipeline for object detection; temperature, altitude, and depth measurements; and the robot's pose, velocity, and acceleration.}
    \label{fig:GA}
\end{figure}

This paper presents SubPipe, a submarine pipeline inspection dataset with ground truth for object detection, image segmentation, and visual-inertial localization. It includes RGB images and side-scan sonar (SSS) images of the seafloor, accelerations captured by an \gls{INS}, as well as the linear velocities estimated by a \gls{DVL}.  Annotations for semantic segmentation are provided for the RGB images from a designated camera, while the side-scan sonar images are annotated for object detection purposes.
Other sensor measurements include forward-looking echo sounder, temperature, altitude, pressure, and depth. The outline of the data-gathering process can be seen in \cref{fig:GA}.

\begin{figure*}
\centering
\Huge
\resizebox{\textwidth}{!}{\begin{tabular}
{c@{\hspace{.7mm}}c@{\hspace{3mm}}c@{\hspace{3mm}}c@{\hspace{3mm}}c@{\hspace{3mm}}c@{\hspace{.7mm}}c@{\hspace{3mm}}c@{\hspace{.7mm}}c@{\hspace{3mm}}c@{\hspace{.7mm}}c}

 \multicolumn{2}{c}{SubPipe} & Aqualoc\cite{rw:dataset:aqualoc} &  AURORA\cite{rw:dataset:bernardi2022aurora} & Caves\cite{rw:dataset:caves} & \multicolumn{2}{c}{MIMIR \cite{rw:dataset:mimir}} & \multicolumn{2}{c}{SUIM\cite{rw:dataset:suim}} & \multicolumn{2}{c}{TrashCan \cite{rw:dataset:trashcan}} \\ 
 
 \adjustbox{valign=m,vspace=.2pt}{\includegraphics[height=.2\linewidth]{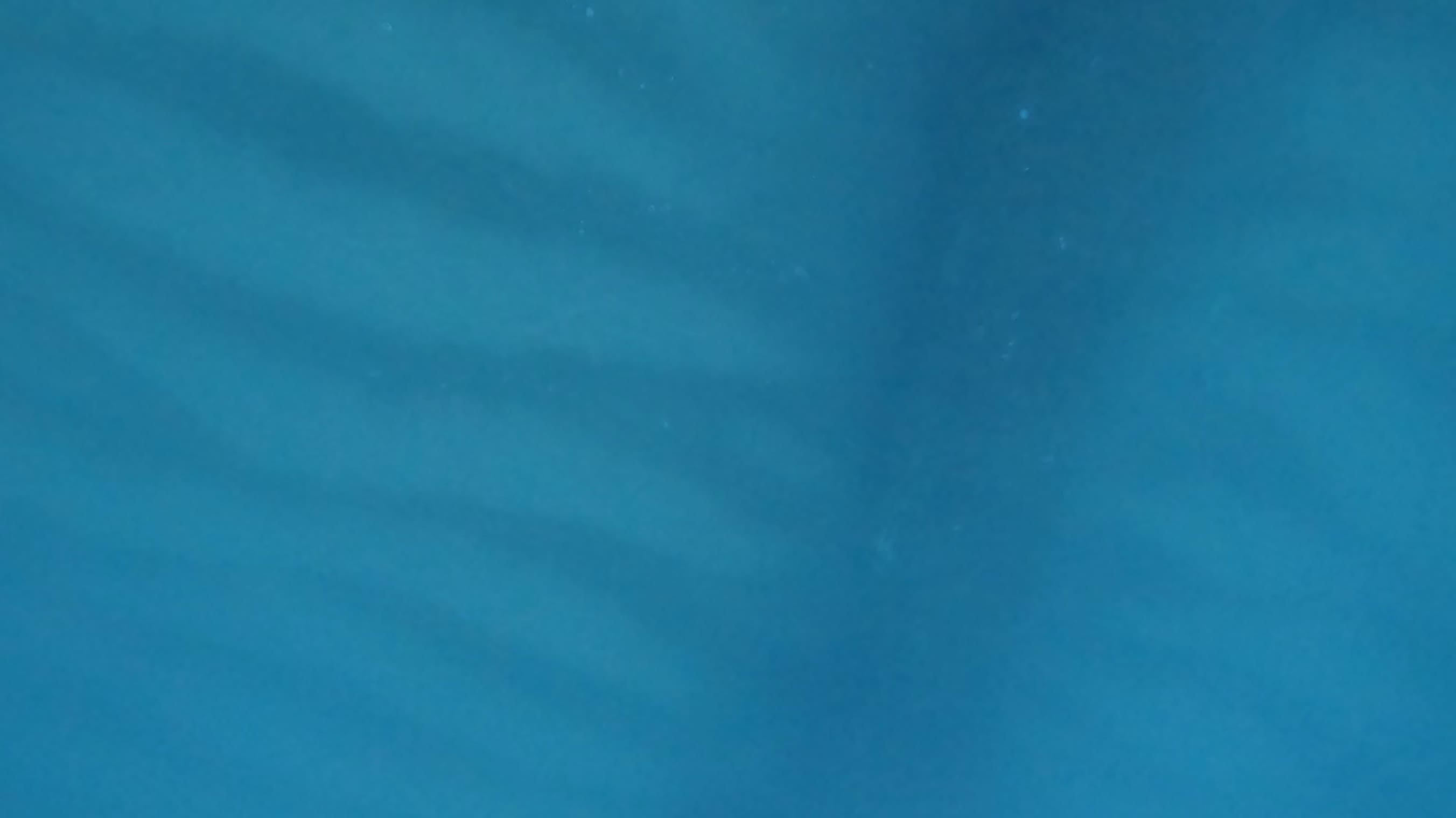}} 
& \adjustbox{valign=m,vspace=.2pt}{\includegraphics[height=.2\linewidth]{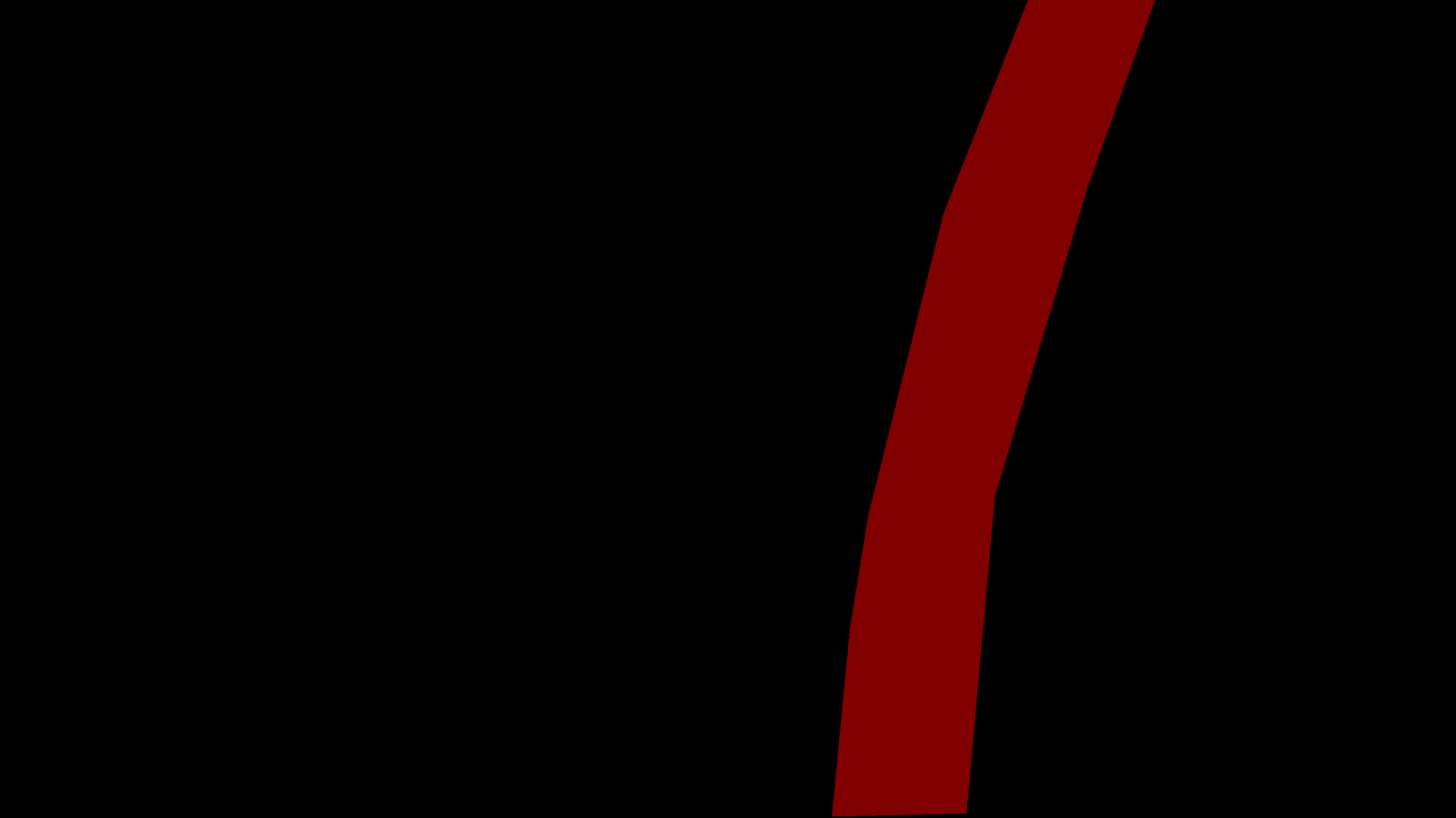}}

& \adjustbox{valign=m,vspace=.2pt}{\includegraphics[height=.2\linewidth]{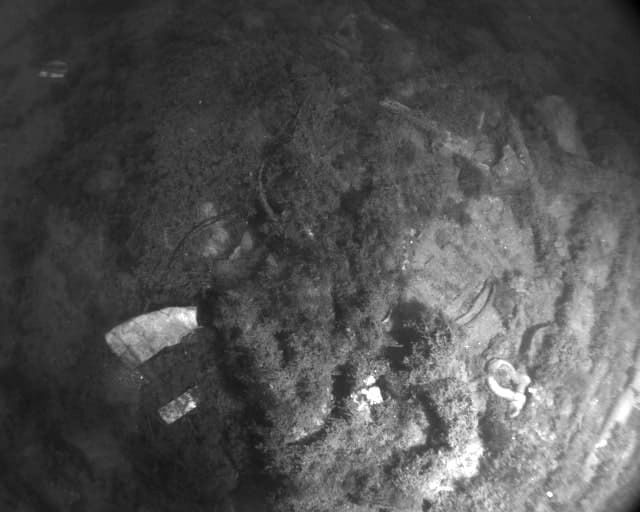}}

& \adjustbox{valign=m,vspace=.2pt}{\includegraphics[height=.2\linewidth]{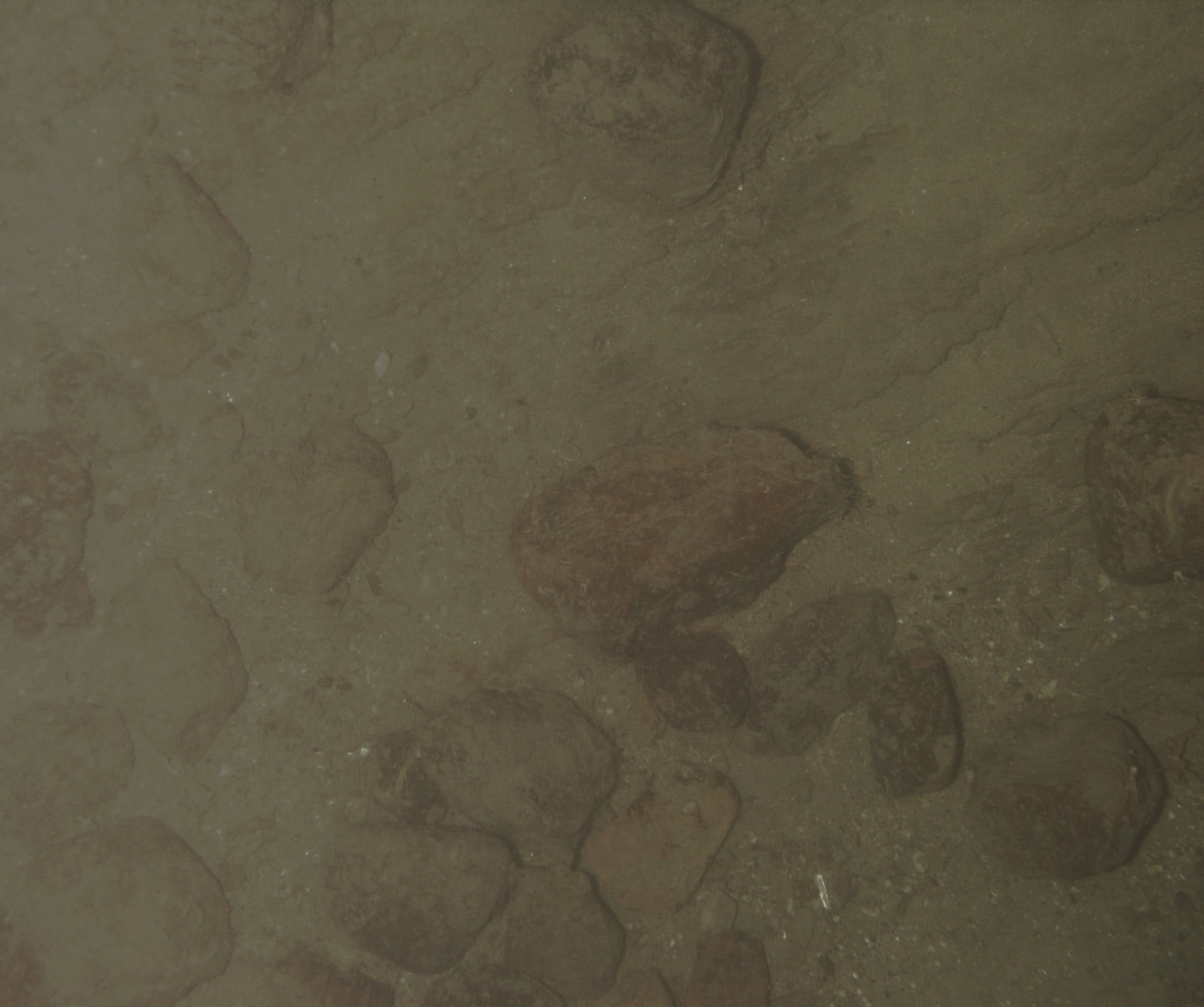}}

& \adjustbox{valign=m,vspace=.2pt}{\includegraphics[height=.2\linewidth]{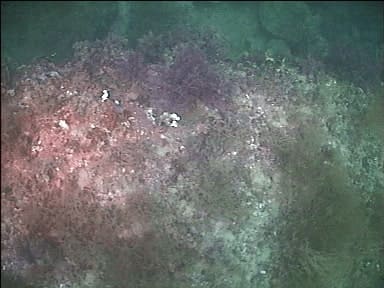}}
& \adjustbox{valign=m,vspace=.2pt}{\includegraphics[height=.2\linewidth]{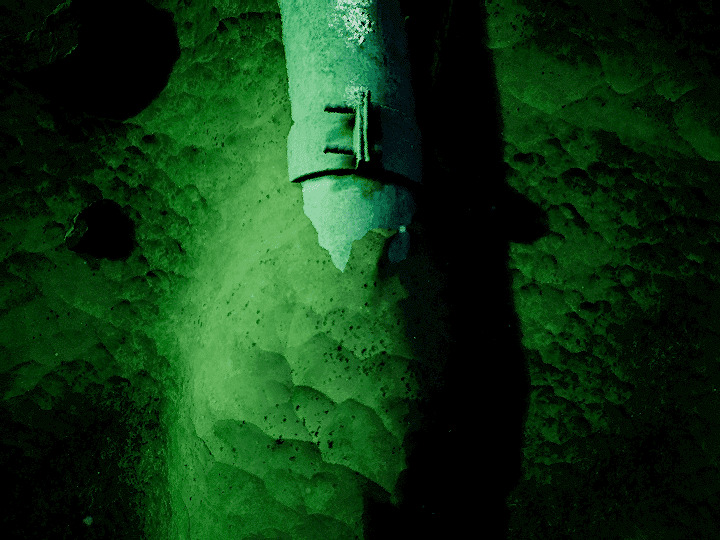}}
& \adjustbox{valign=m,vspace=.2pt}{\includegraphics[height=.2\linewidth]{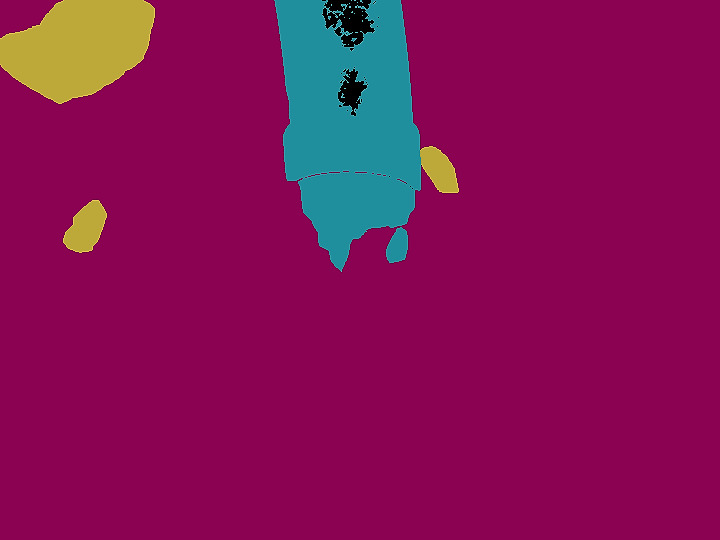}}
& \adjustbox{valign=m,vspace=.2pt}{\includegraphics[height=.2\linewidth]{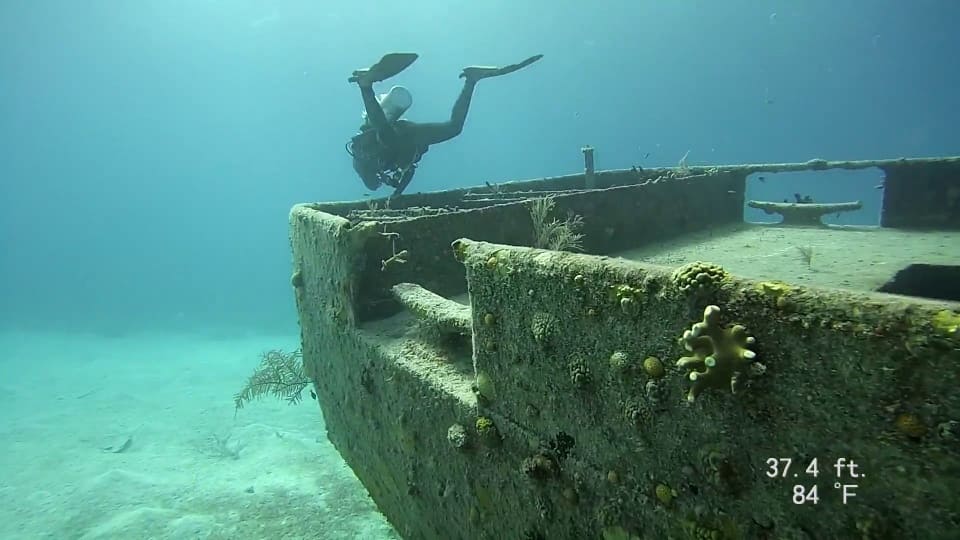}} 
& \adjustbox{valign=m,vspace=.2pt}{\includegraphics[height=.2\linewidth]{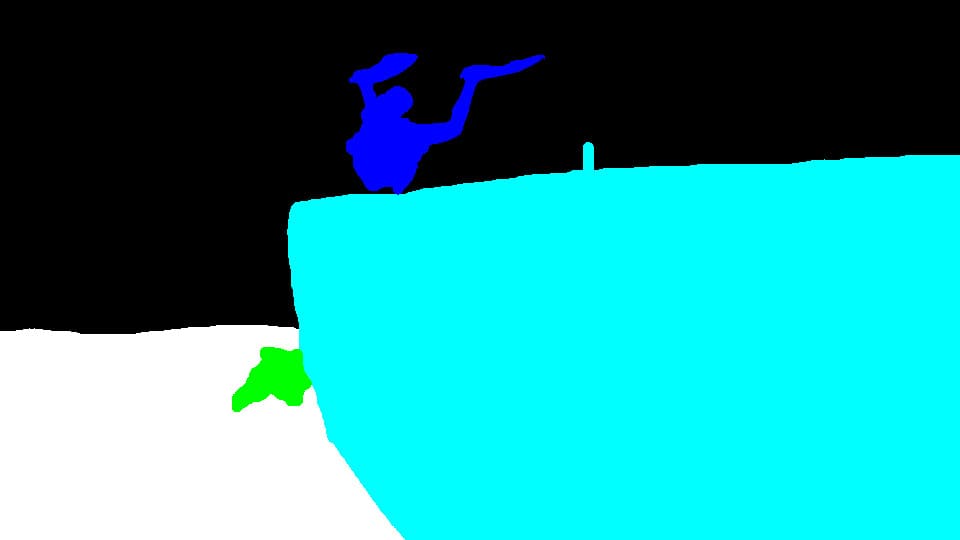}} 
& \adjustbox{valign=m,vspace=.2pt}{\includegraphics[height=.2\linewidth]{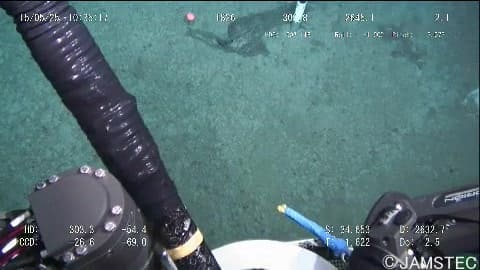}} 
& \adjustbox{valign=m,vspace=.2pt}{\includegraphics[height=.2\linewidth]{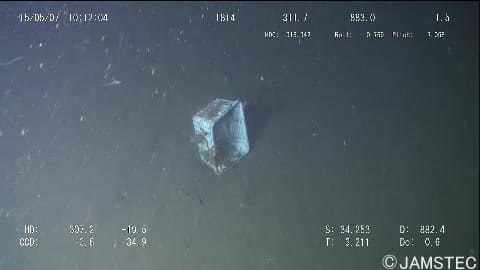}} 
\end{tabular}}
\caption{Sample images from SubPipe and other state-of-the-art underwater datasets. A sample segmentation image is also shown for those datasets that include segmentation labels.}
\label{fig:sampledatasetimgs}
\end{figure*}

As shown in \cref{table:comparisonsoadataset}, the dataset closest to SubPipe is MIMIR-UW \cite{rw:dataset:mimir}, with a similar set of sensors and pipe segmentation labels. However, MIMIR-UW presents a simulated dataset with fewer underwater artifacts, such as scattering and blur.  \Cref{fig:sampledatasetimgs} visually compares the imaging conditions between SubPipe and some state-of-the-art underwater datasets.

\begin{table}[!htbp]
\caption{Comparison of state-of-the-art underwater datasets. \\ "Seg." and "Det." indicates annotations for segmentation and object detection, while "Depth" notes range-based imagery (such as side-scan sonar and depth camera).}
\centering
\footnotesize
\label{table:comparisonsoadataset}
\begin{tabular}{l@{\hspace{.3mm}} c@{\hspace{.8mm}} c@{\hspace{.8mm}} c@{\hspace{.4mm}} c@{\hspace{.3mm}} c@{\hspace{.7mm}} r }
\toprule
Dataset                                      & Pose      & Seg. & Det. & Camera    & Depth   & Object labels\\    
\midrule
Aqualoc \cite{rw:dataset:aqualoc}            & \cmark    & \xmark & \xmark       & mono & \xmark  & N\slash A\\
AURORA \cite{rw:dataset:bernardi2022aurora}  & \cmark    & \xmark & \xmark        & mono & \xmark  & N\slash A\\
MIMIR-UW\cite{rw:dataset:mimir}              & \cmark    & \cmark & \xmark        &rig        & \cmark  & Marine life; pipes\\  
NAUTEC UWI\cite{rw:dataset:nautec}           & \xmark    & \cmark & \xmark       & mono & \xmark  & Marine life; other\\
SUIM\cite{rw:dataset:suim}                   & \xmark    & \cmark  & \xmark      & mono & \xmark  & Marine life; other\\
TrashCan\cite{rw:dataset:trashcan}           & \xmark    & \cmark  & \cmark      & mono & \xmark  & Marine debris\\
\hline
\textbf{SubPipe (ours)}             & \cmark    & \cmark   & \cmark  &  mono(x2)       & \cmark  & Pipe \\    
\bottomrule

\end{tabular}
\end{table}

By releasing SubPipe, we aim to provide the research community with a novel and multimodal underwater dataset, facilitating advancements in underwater computer vision algorithms. To underscore the dataset's significance, we propose a series of experiments that demonstrate the necessity of SubPipe and showcase the novel challenges that state-of-the-art algorithms face in underwater scenarios. Through this contribution, we envision fostering a collaborative effort toward developing robust and versatile underwater computer vision solutions.

The rest of this paper is structured as follows. Section \ref{sec:dataset} presents SubPipe, including information about the sensors and techniques used for gathering the dataset. Furthermore, Section \ref{section:metrics} provides insights about the dataset's quality by analyzing different metrics and comparing it with other state-of-the-art datasets. Section \ref{sec:evaluations} lists the conducted experiments and their results. Finally, Section \ref{sec:conclusion} concludes this paper.

\section{The Dataset}
\label{sec:dataset}

\renewcommand{\arraystretch}{1.5}
\begin{table*}
    \centering
    \small
    \begin{tabular}{lp{.12\textwidth}cc@{\hspace{0.3mm}}c@{\hspace{0.3mm}}c}
    \toprule
        \multicolumn{2}{l}{Chunk}  & Trajectory [m] / [$\deg$] & Cam0 & Cam1 & Side-scan  \\
\midrule
        0    & Time: 9m 0.54s Length: 571.7m \#Poses: 16199     
        &  \adjustbox{valign=m,vspace=.1mm}{\includegraphics[height=0.1\textwidth]{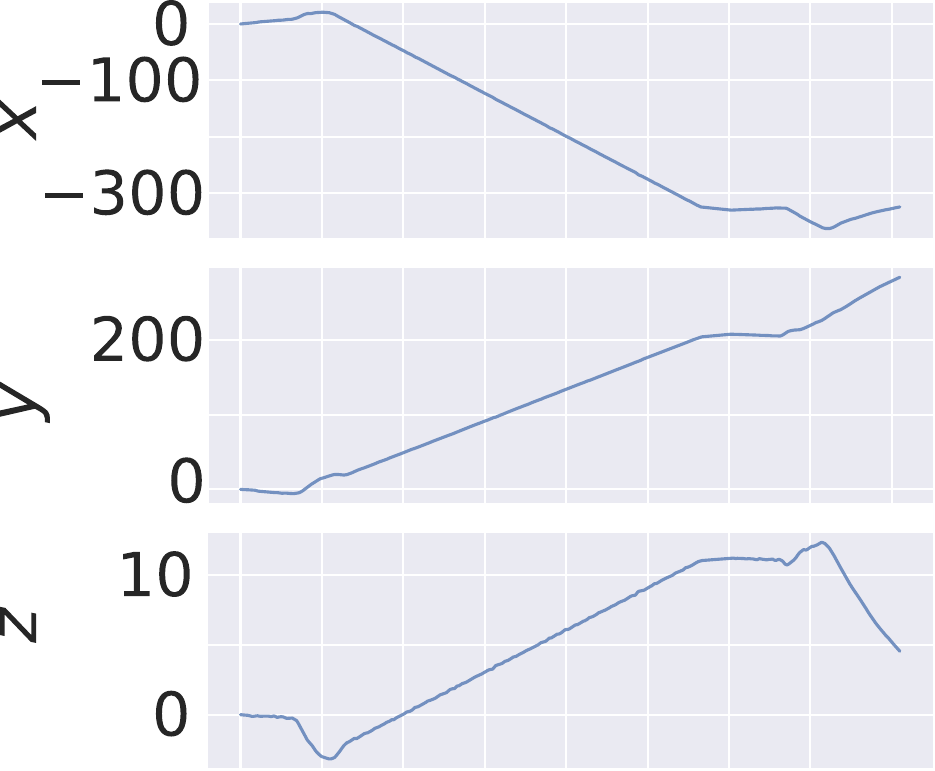}} 
         \adjustbox{valign=m,vspace=.1mm}{\includegraphics[height=0.1\textwidth]{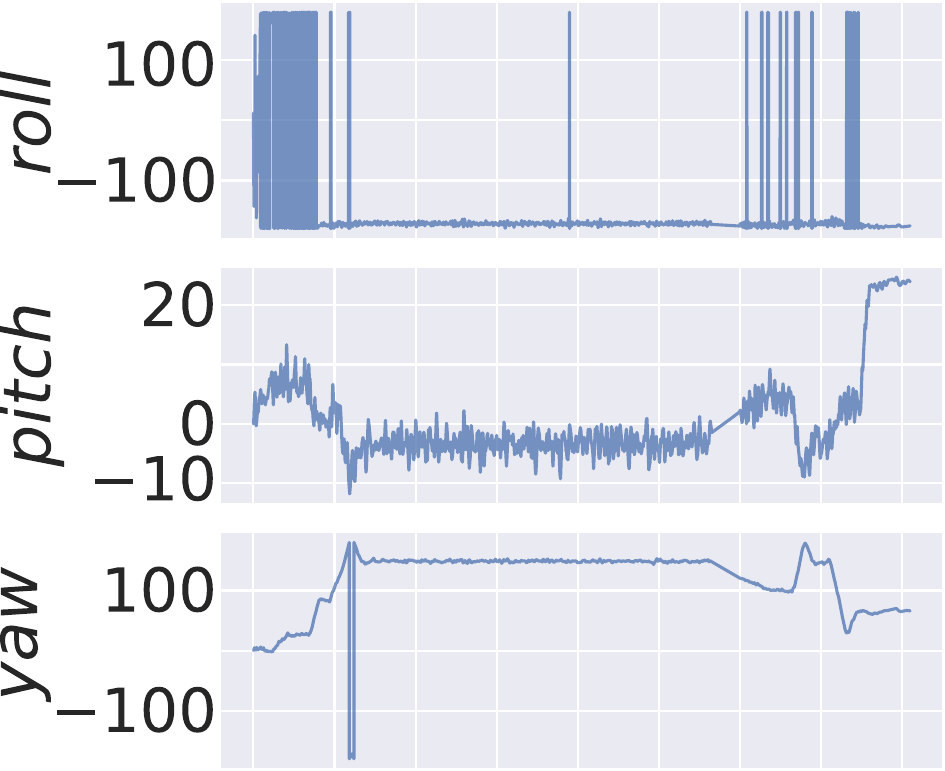}}  
        & \adjustbox{valign=m,vspace=.1mm}{\includegraphics[height=0.1\textwidth]{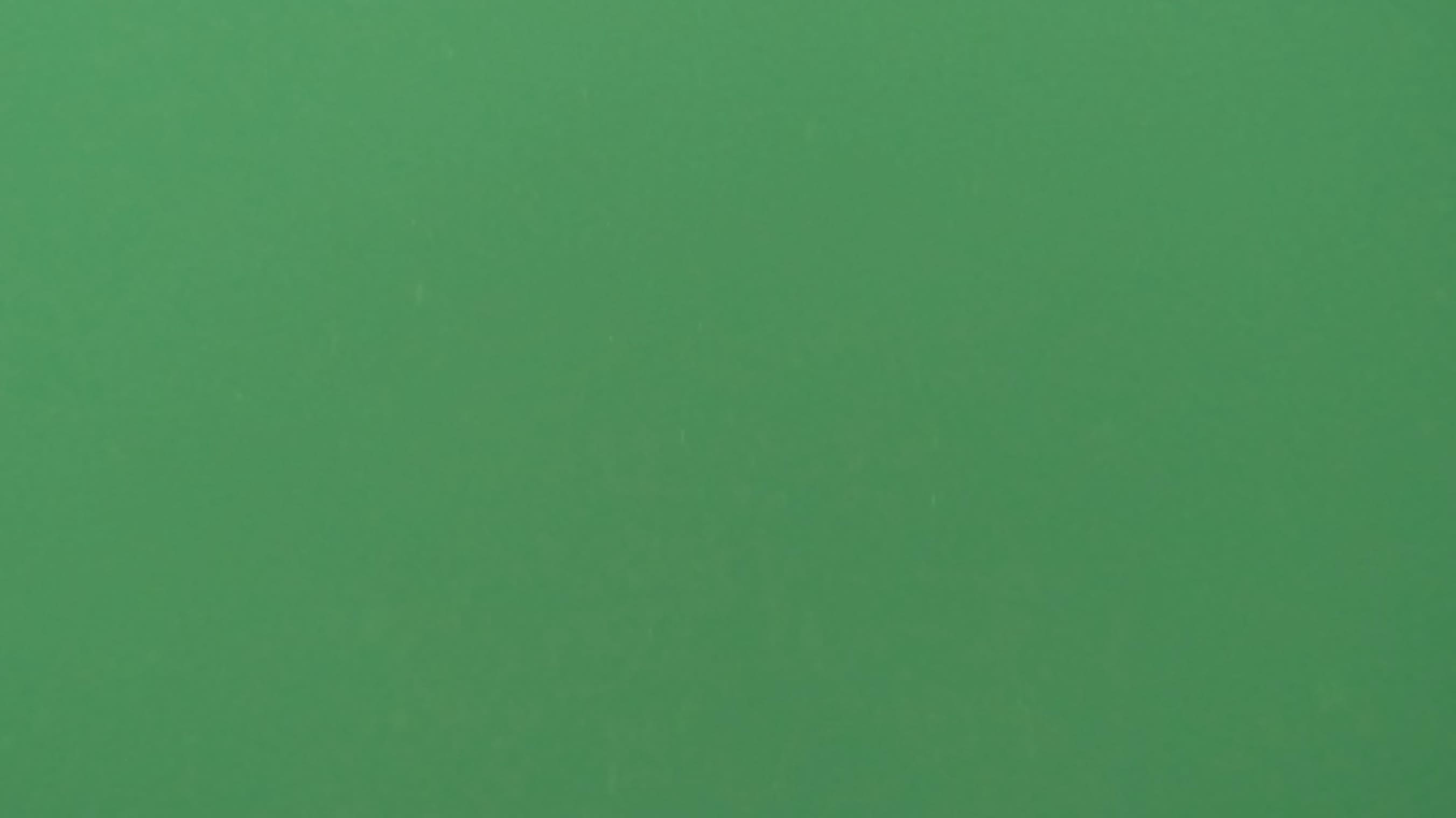}} 
        & \adjustbox{valign=m,vspace=.1mm}{\includegraphics[height=0.1\textwidth]{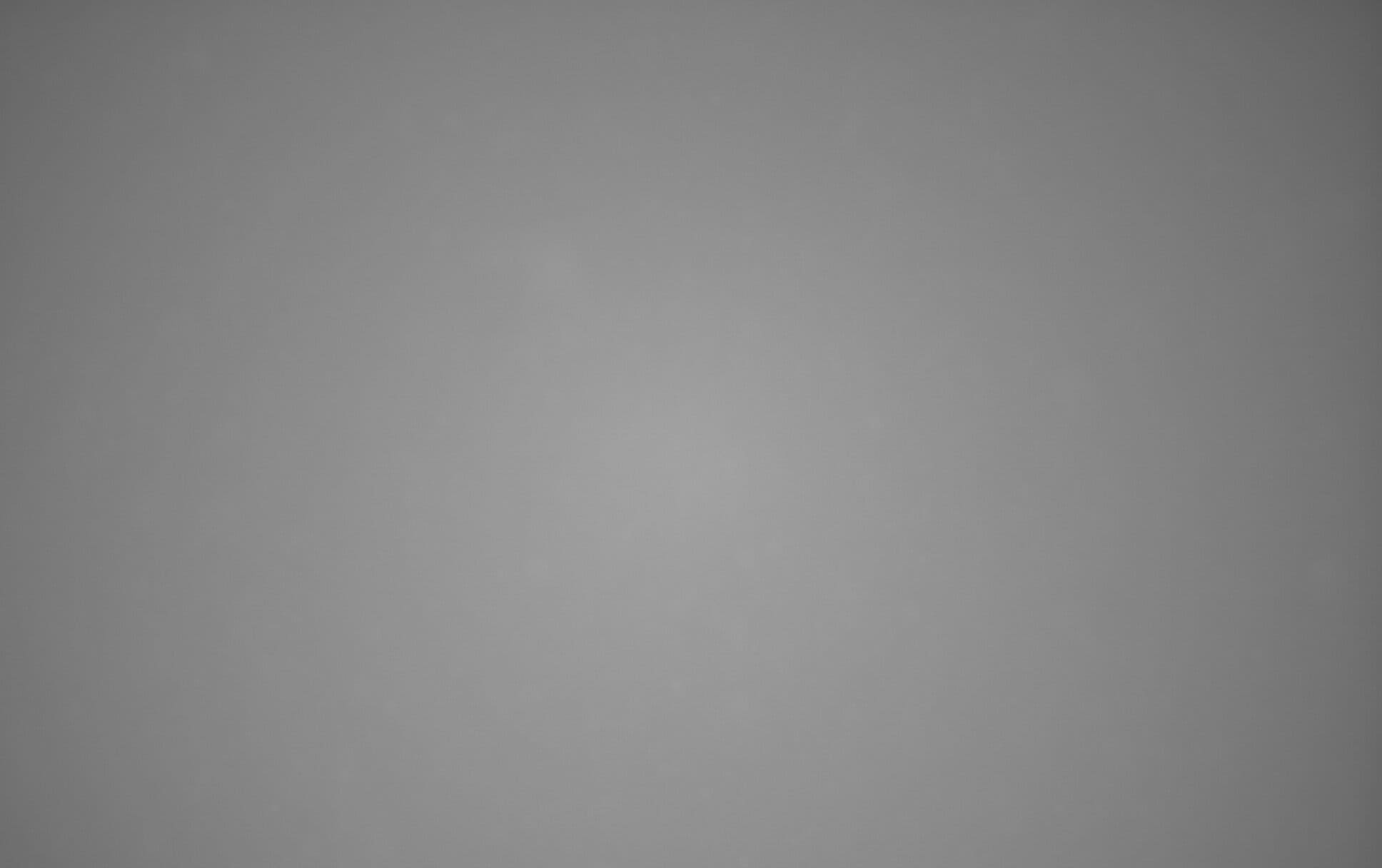}} 
        & \adjustbox{valign=m,vspace=.1mm}{\includegraphics[height=0.1\textwidth, width=0.2\textwidth ]{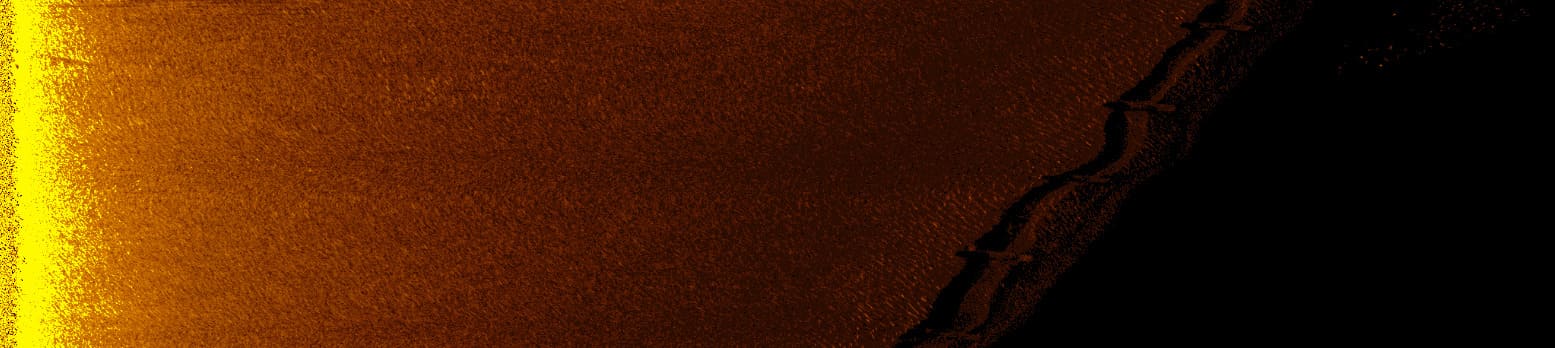}} \\ 
        1    & Time 9m 0.54s Length: 466.8m \#Poses: 16199      &  \adjustbox{valign=m,vspace=.15mm}{\includegraphics[height=0.1\textwidth]{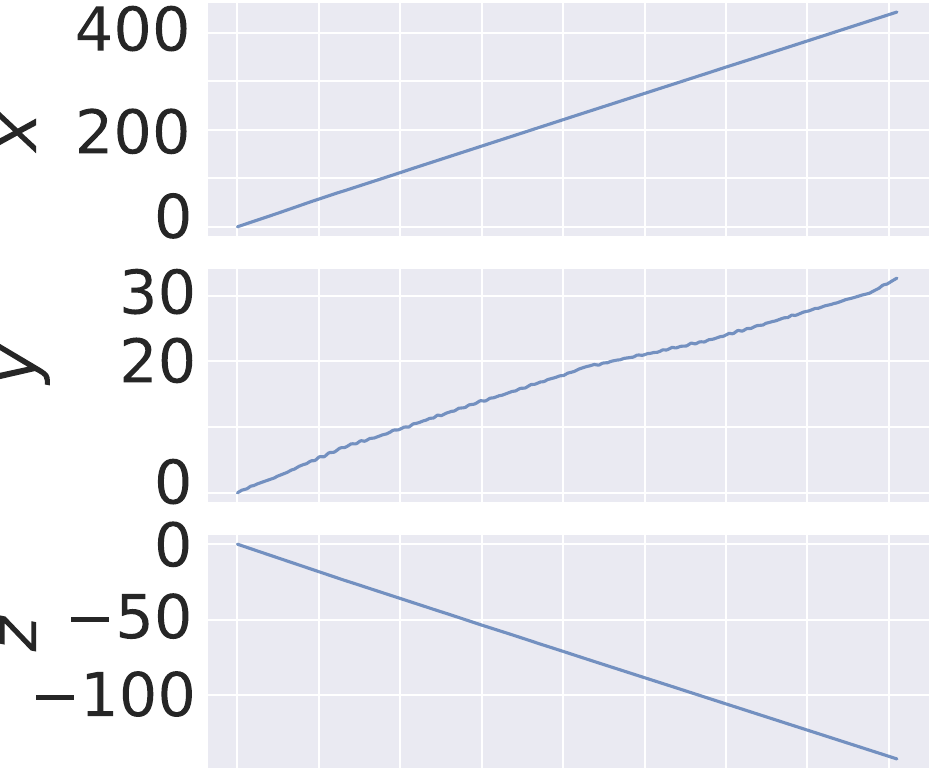}} \adjustbox{valign=m,vspace=.15mm}{\includegraphics[height=0.1\textwidth]{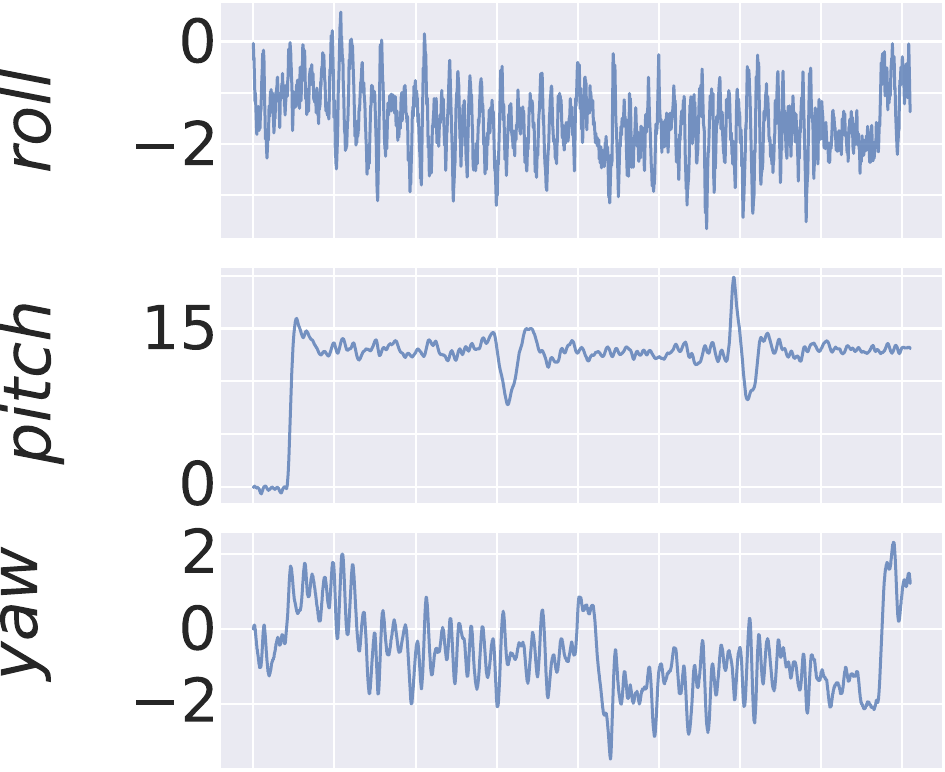}} 
        & \adjustbox{valign=m,vspace=.15mm}{\includegraphics[height=0.1\textwidth]{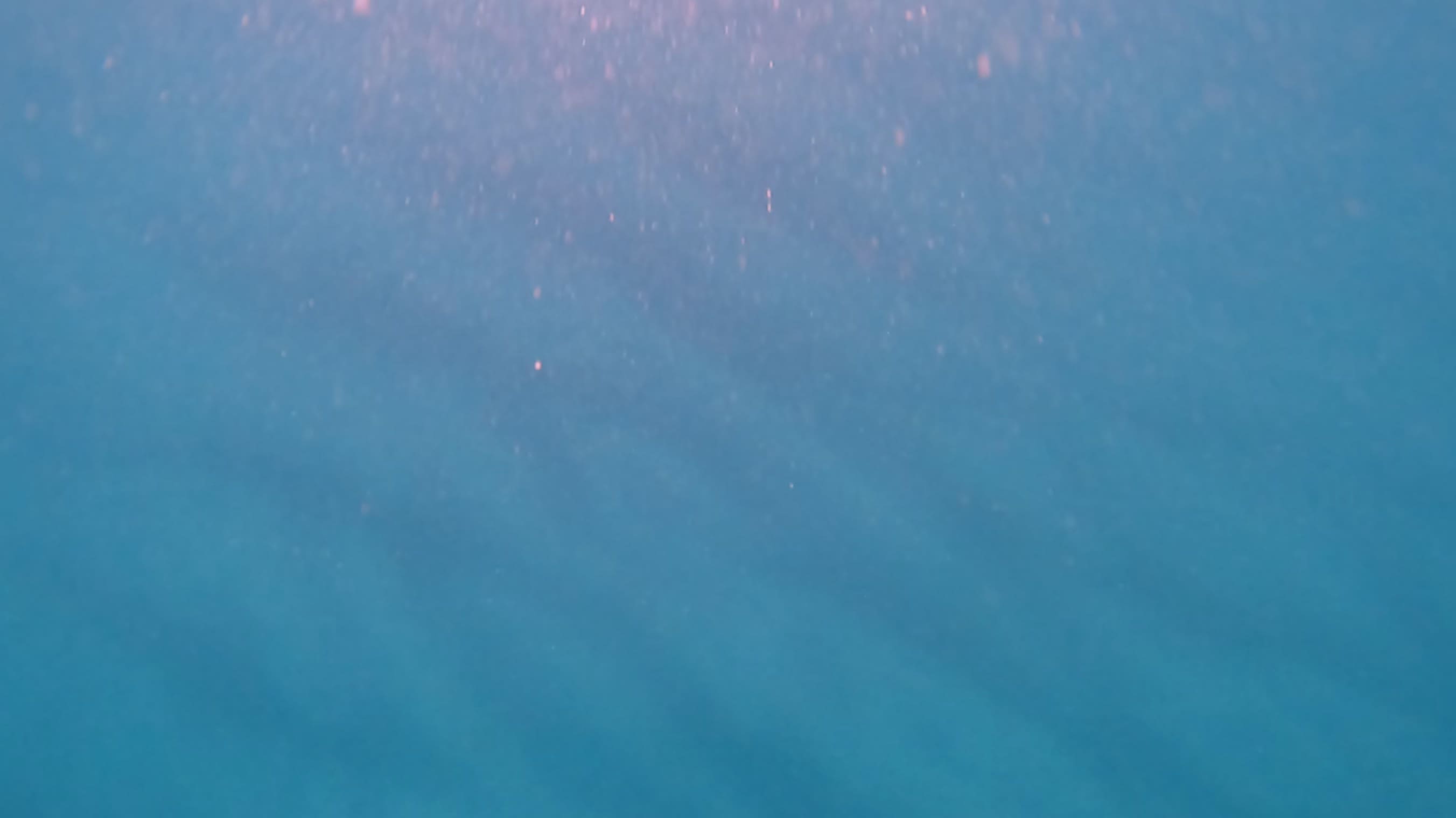}} 
        & \adjustbox{valign=m,vspace=.15mm}{\includegraphics[height=0.1\textwidth]{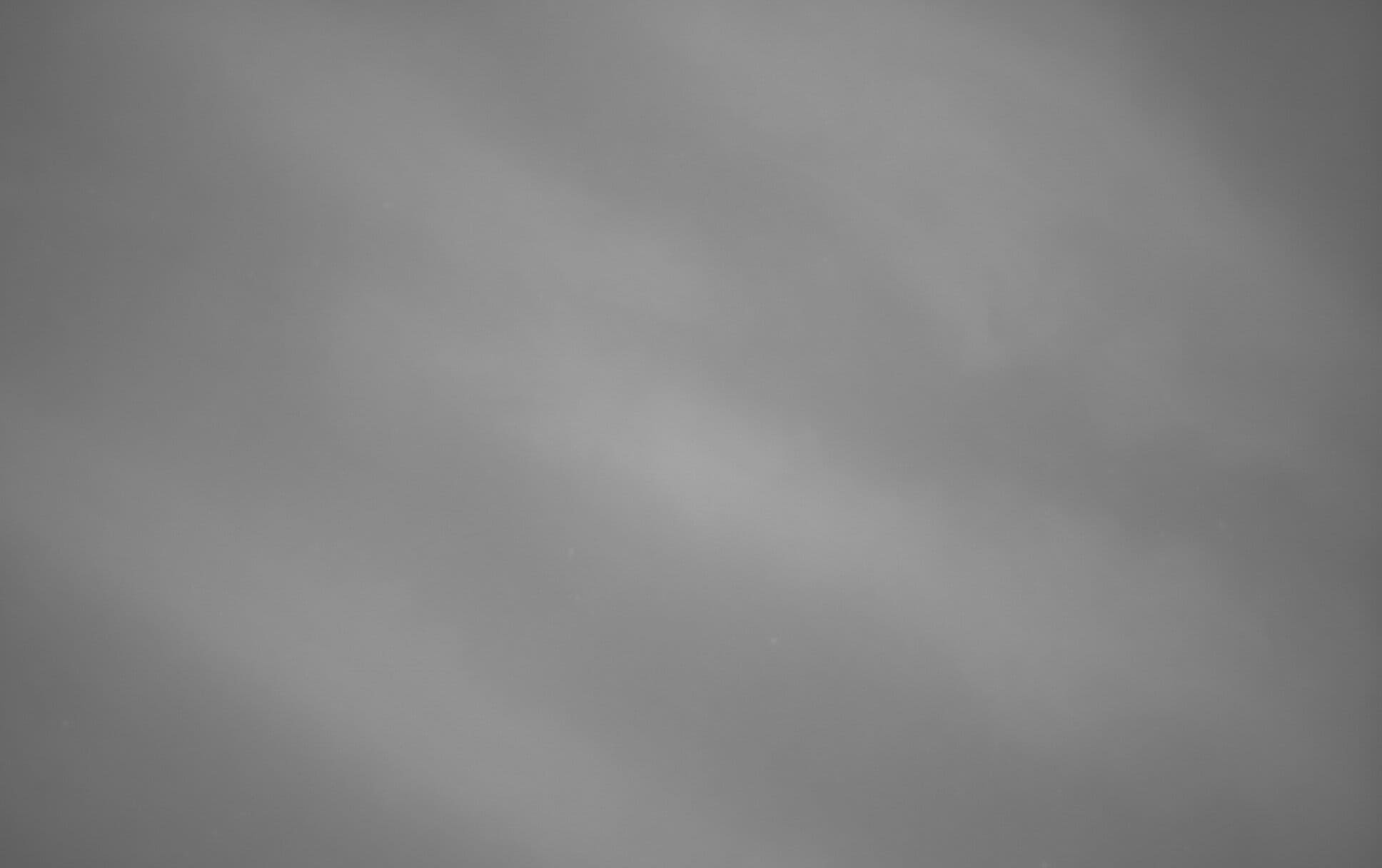}} 
        & \adjustbox{valign=m,vspace=.15mm}{\includegraphics[height=0.1\textwidth, width=0.2\textwidth ]{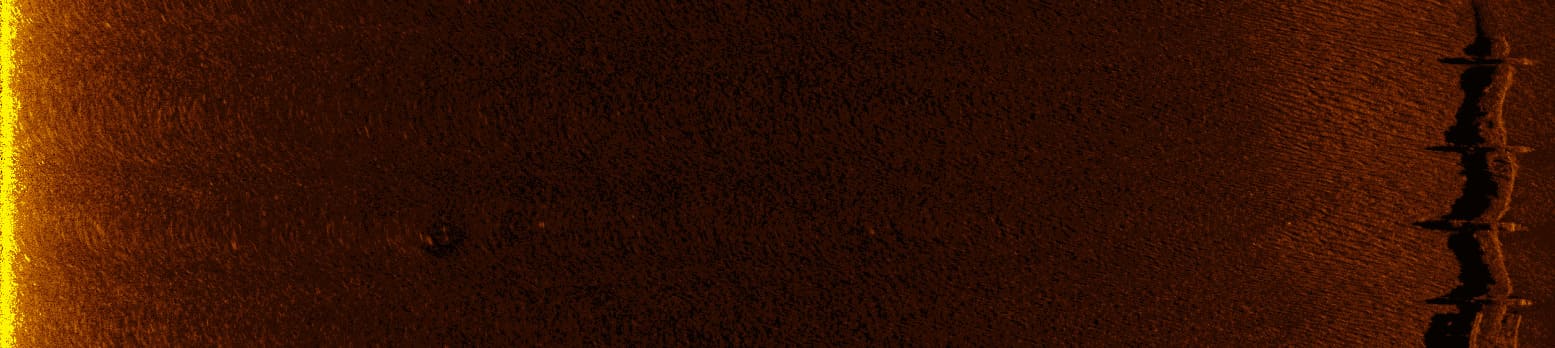}} \\
        2    &  Time: 8m 59.54s Length: 463.4m \#Poses: 16169      &  \adjustbox{valign=m,vspace=.15mm}{\includegraphics[height=0.1\textwidth]{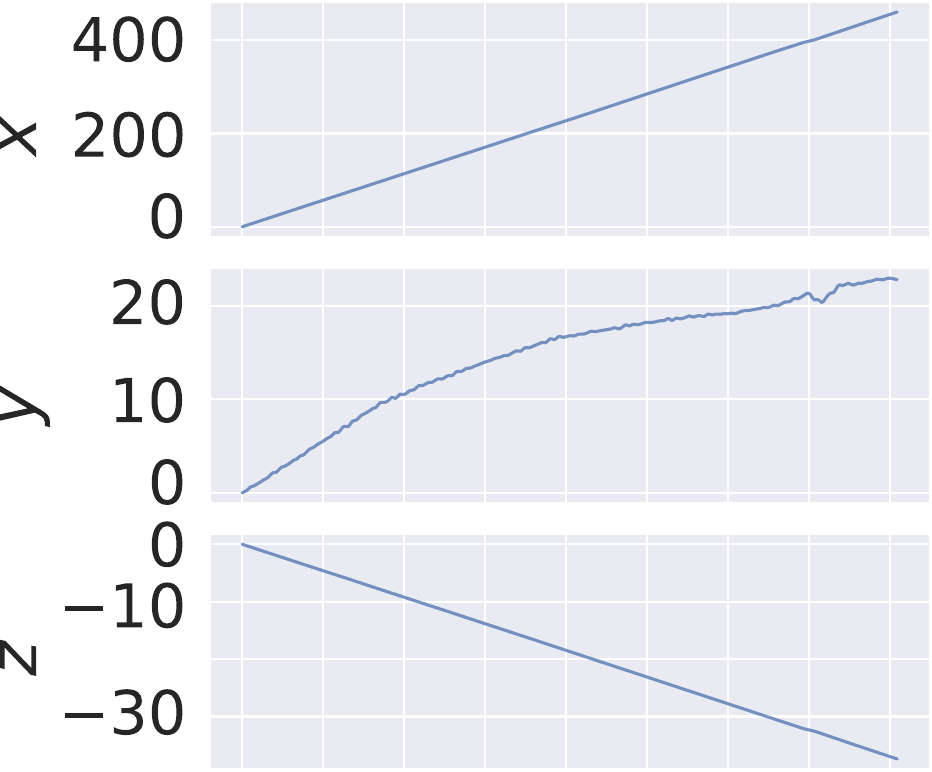}} \adjustbox{valign=m,vspace=.15mm}{\includegraphics[height=0.1\textwidth]{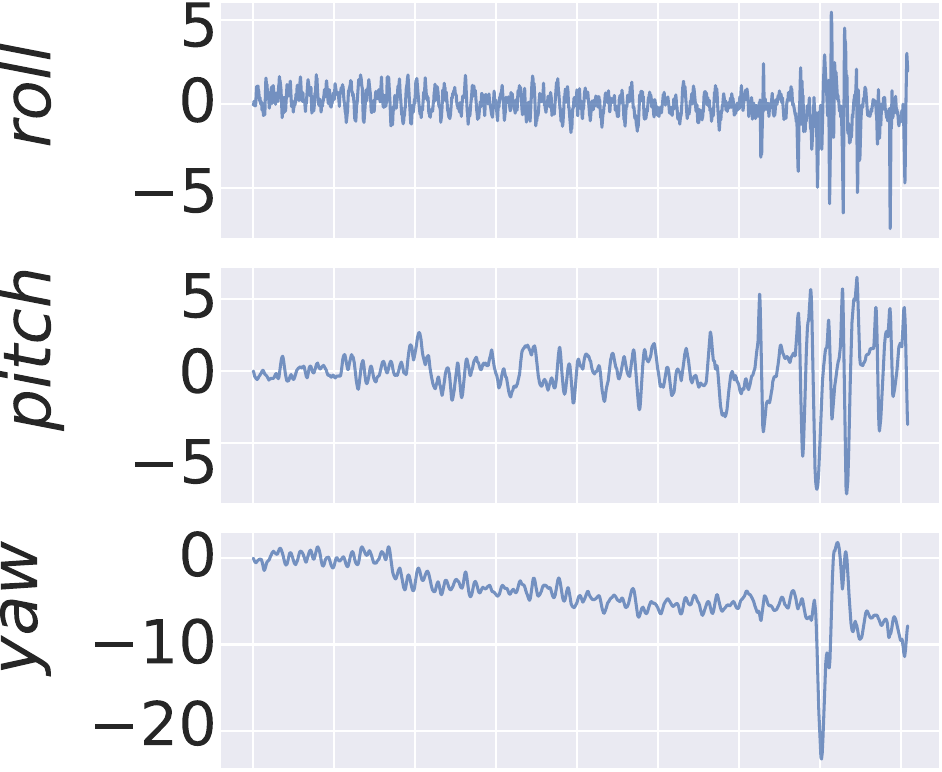}}
        & \adjustbox{valign=m,vspace=.15mm}{\includegraphics[height=0.1\textwidth]{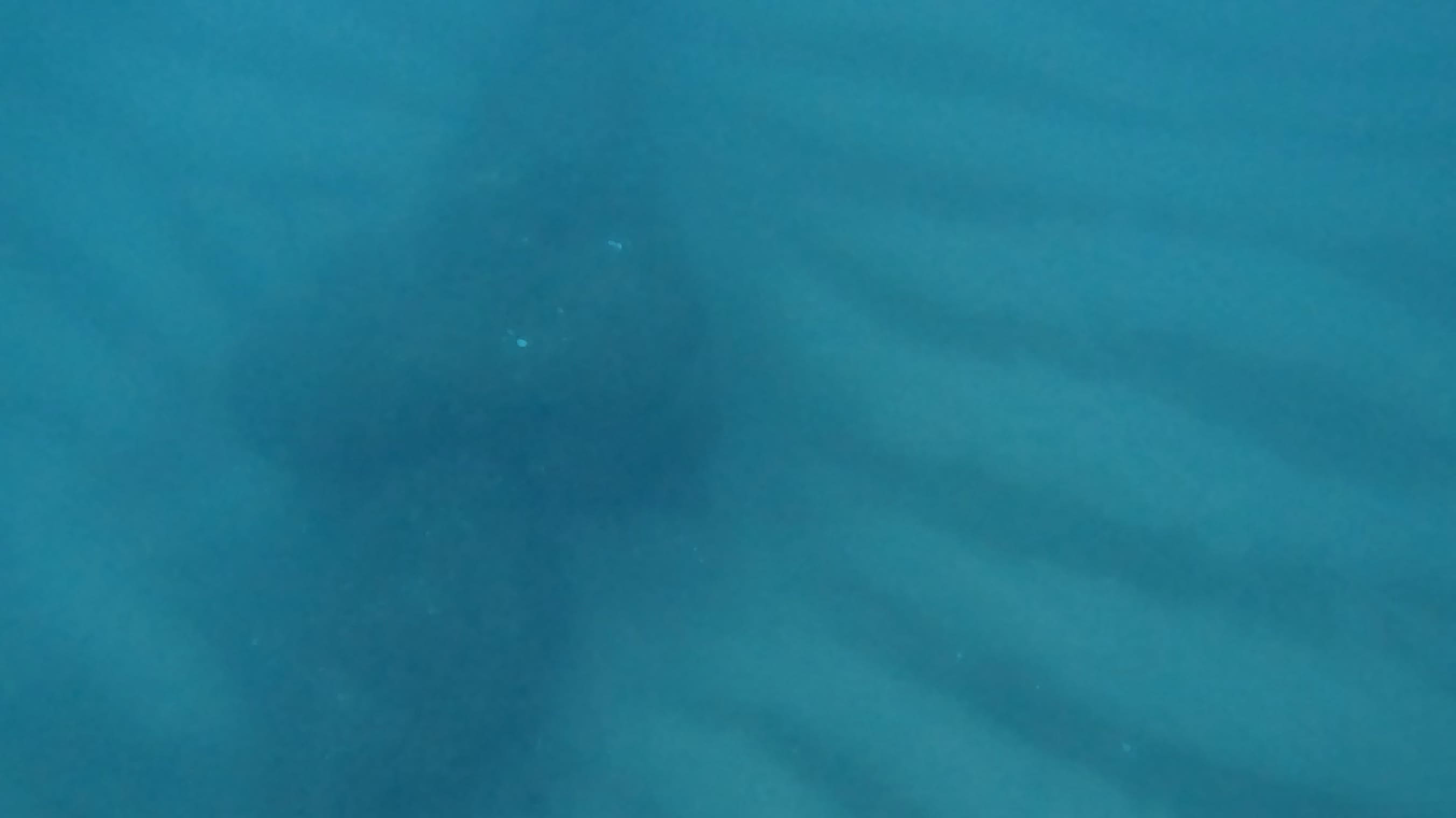}} 
        & \adjustbox{valign=m,vspace=.15mm}{\includegraphics[height=0.1\textwidth]{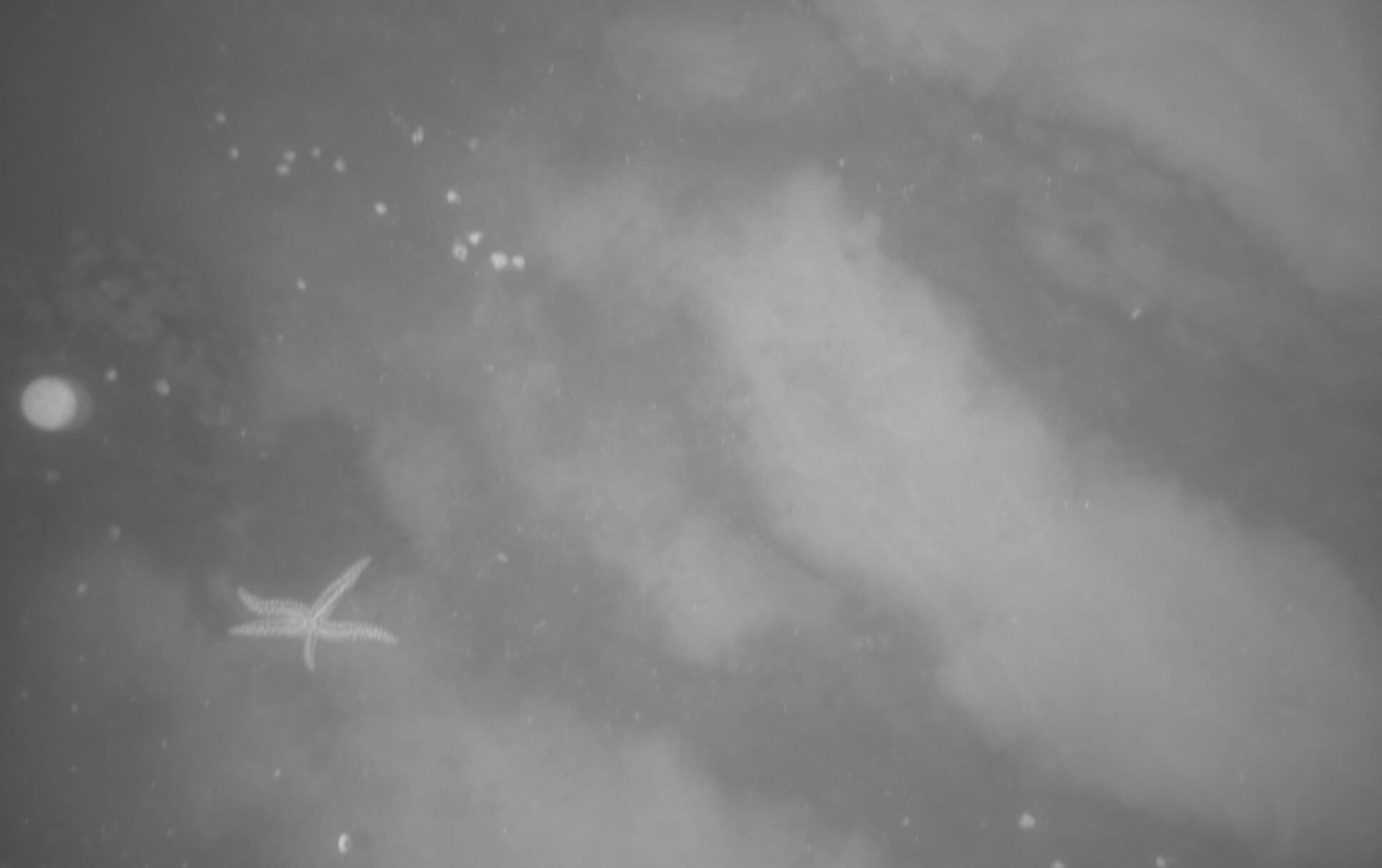}} 
        & \adjustbox{valign=m,vspace=.15mm}{\includegraphics[height=0.1\textwidth, width=0.2\textwidth]{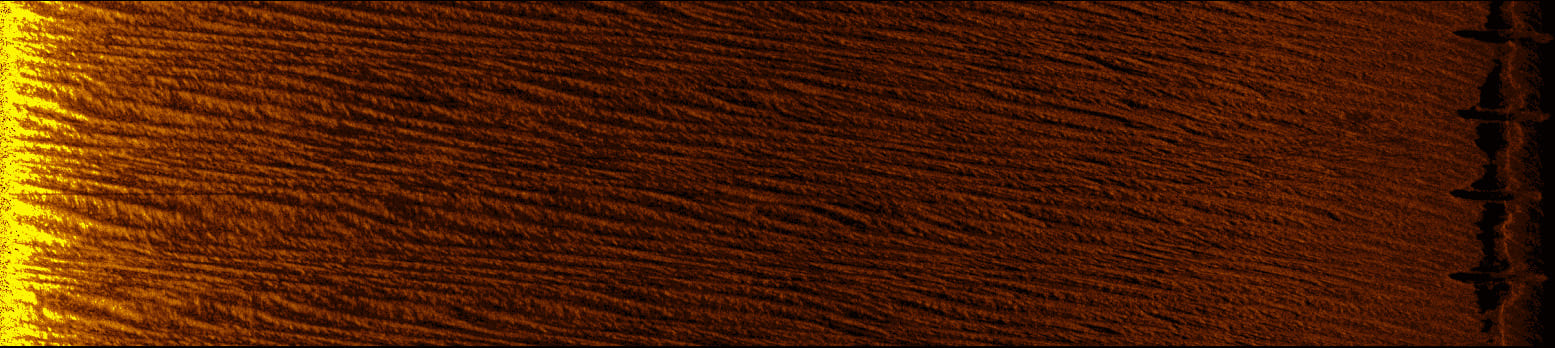}} \\
        3    &  Time: 8m 59.54s Length: 372.7m \#Poses: 16169      &  \adjustbox{valign=m,vspace=.15mm}{\includegraphics[height=0.1\textwidth]{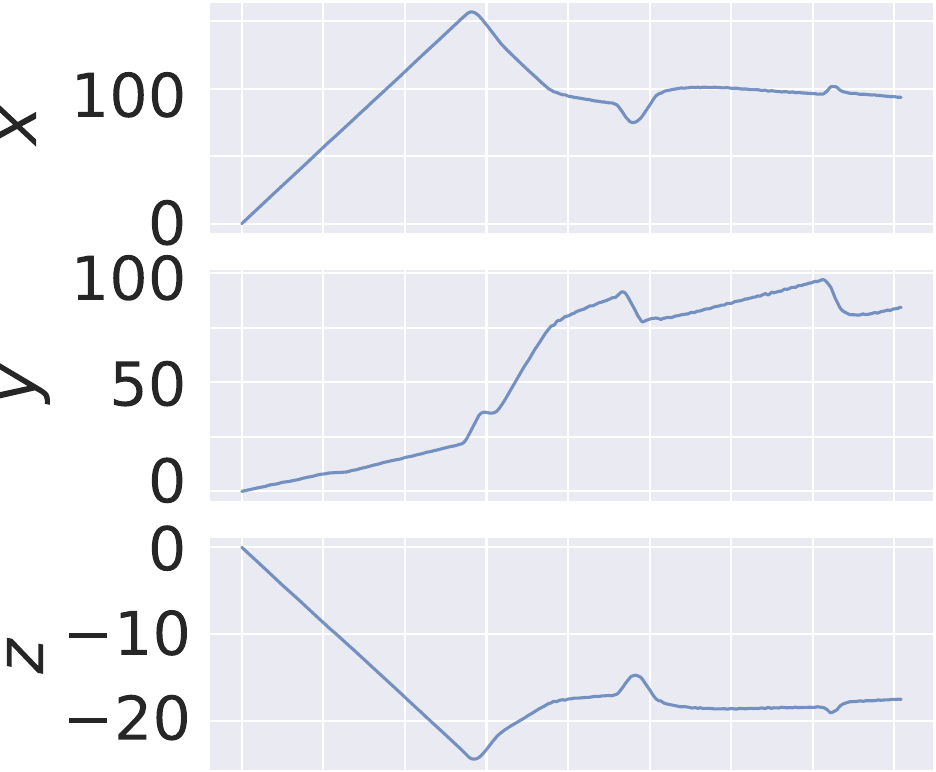}} \adjustbox{valign=m,vspace=.15mm}{\includegraphics[height=0.1\textwidth]{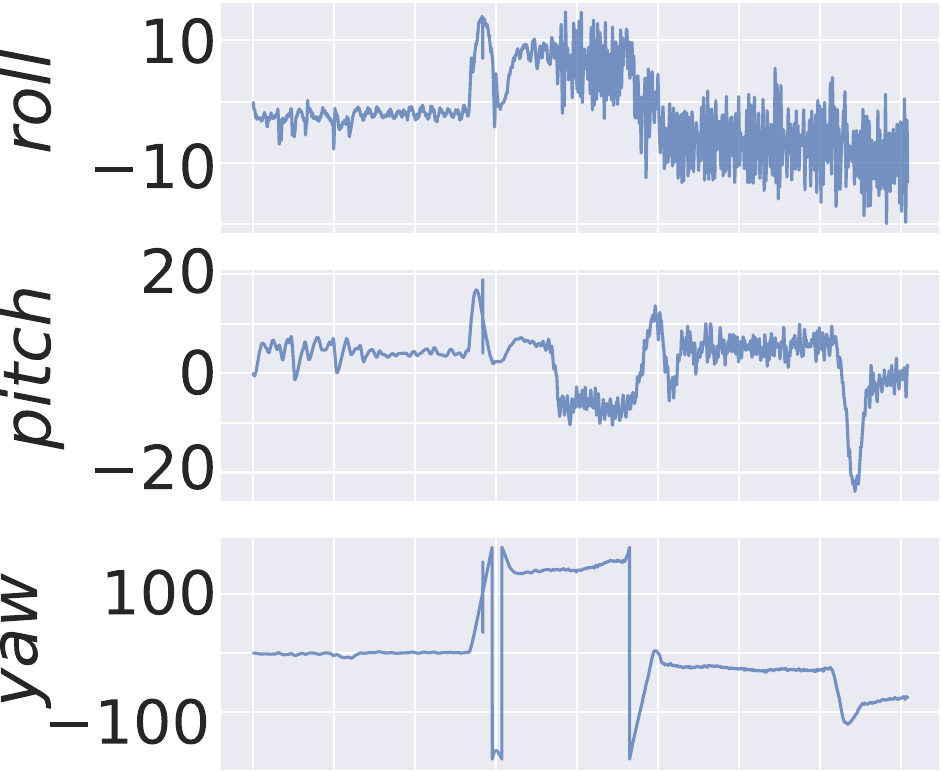}} 
        & \adjustbox{valign=m,vspace=.15mm}{\includegraphics[height=0.1\textwidth]{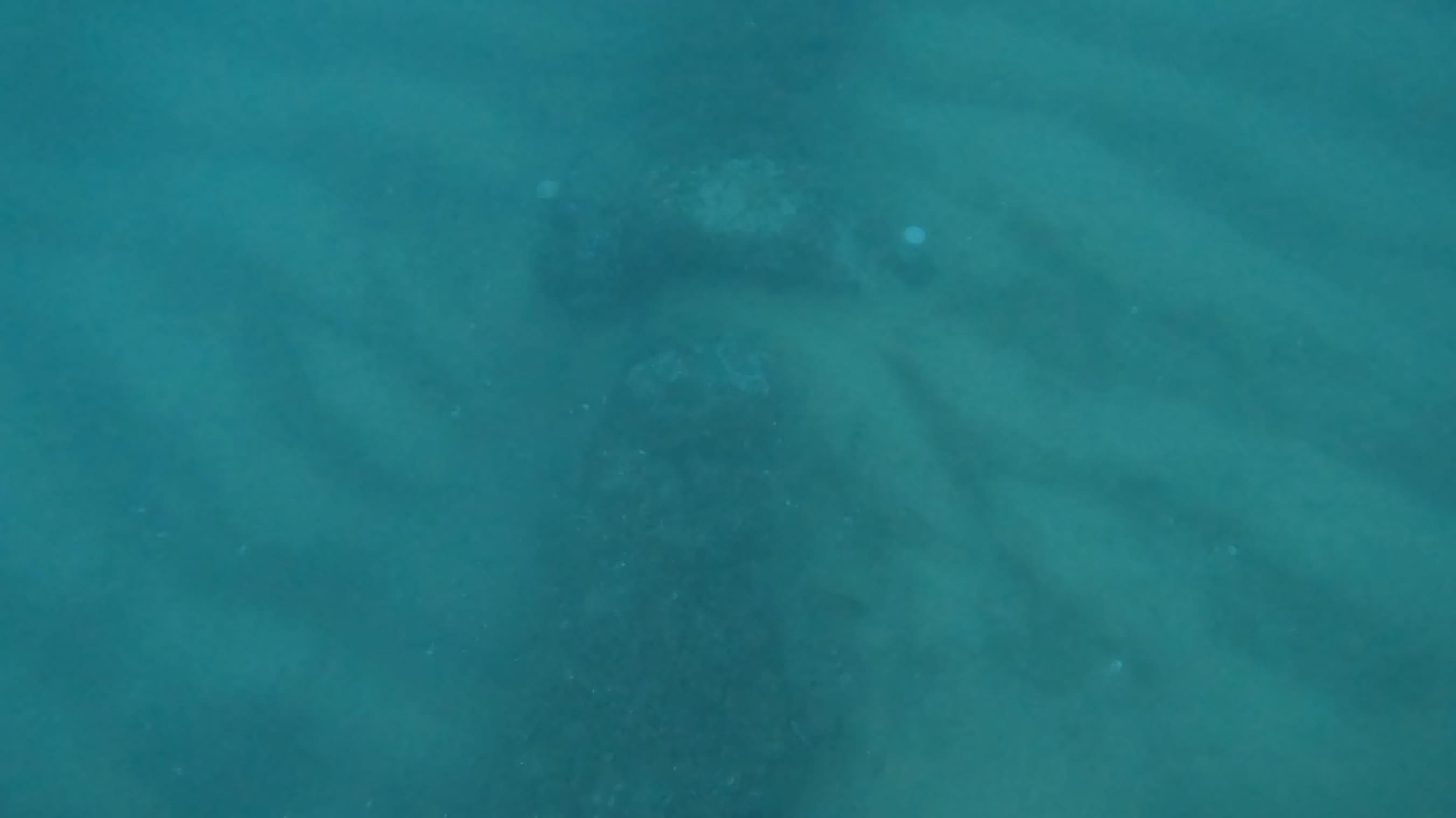}} 
        & \adjustbox{valign=m,vspace=.15mm}{\includegraphics[height=0.1\textwidth]{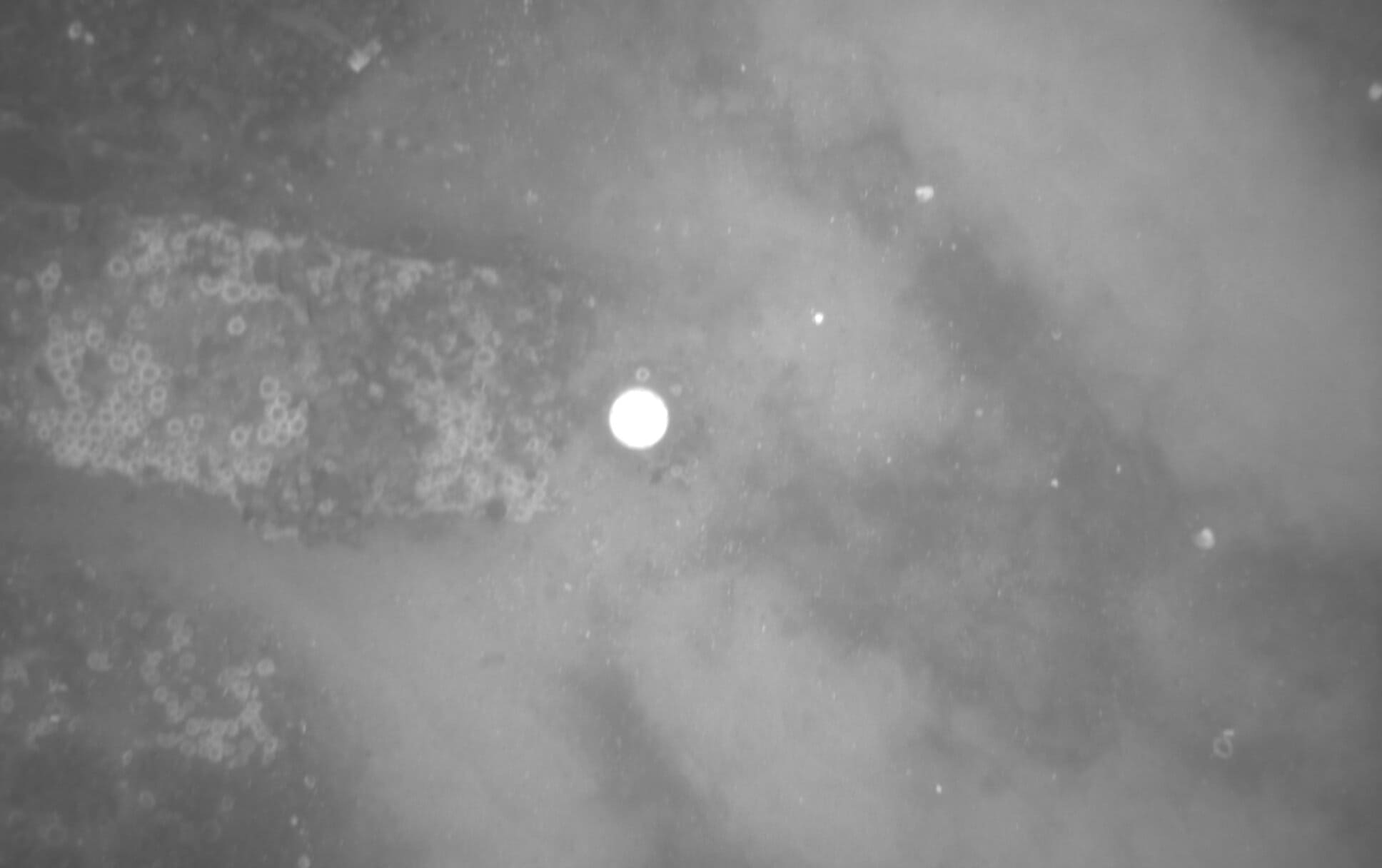}} 
        & \adjustbox{valign=m,vspace=.15mm}{\includegraphics[height=0.1\textwidth, width=0.2\textwidth]{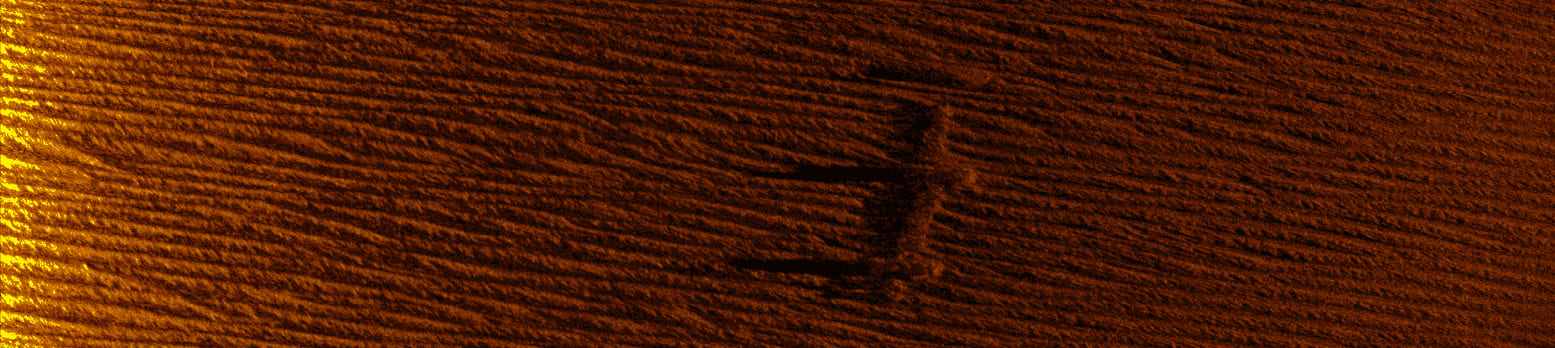}} \\ 
        4     &  Time: 7m 15.73s Length: 374.1m \#Poses: 13057      &  \adjustbox{valign=m,vspace=.15mm}{\includegraphics[height=0.1\textwidth]{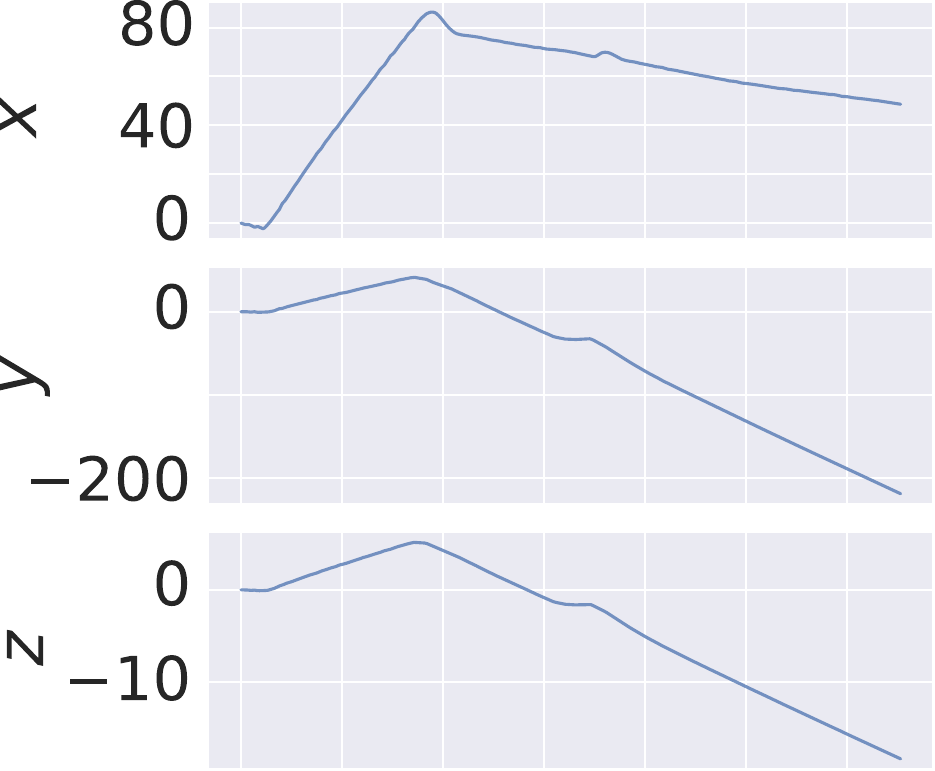}} \adjustbox{valign=m,vspace=.15mm}{\includegraphics[height=0.1\textwidth]{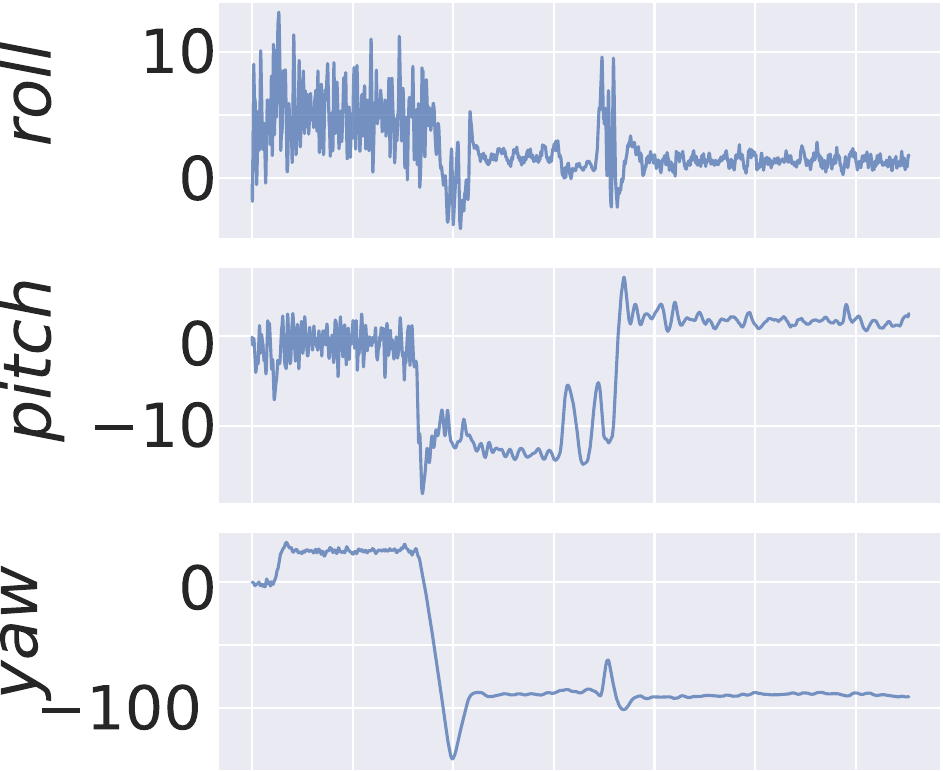}} 
        & \adjustbox{valign=m,vspace=.15mm}{\includegraphics[height=0.1\textwidth]{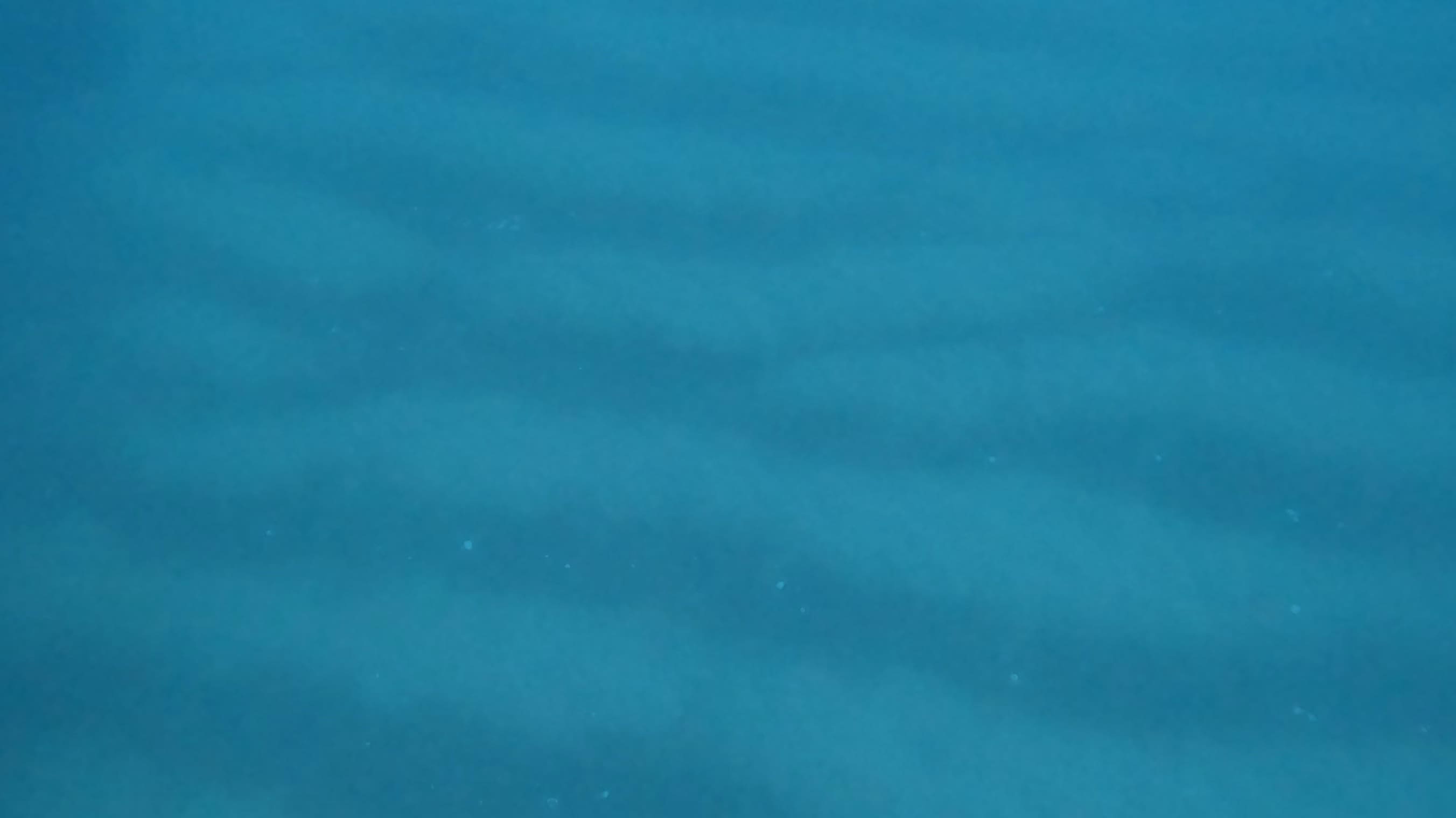}} 
        & \adjustbox{valign=m,vspace=.15mm}{\includegraphics[height=0.1\textwidth]{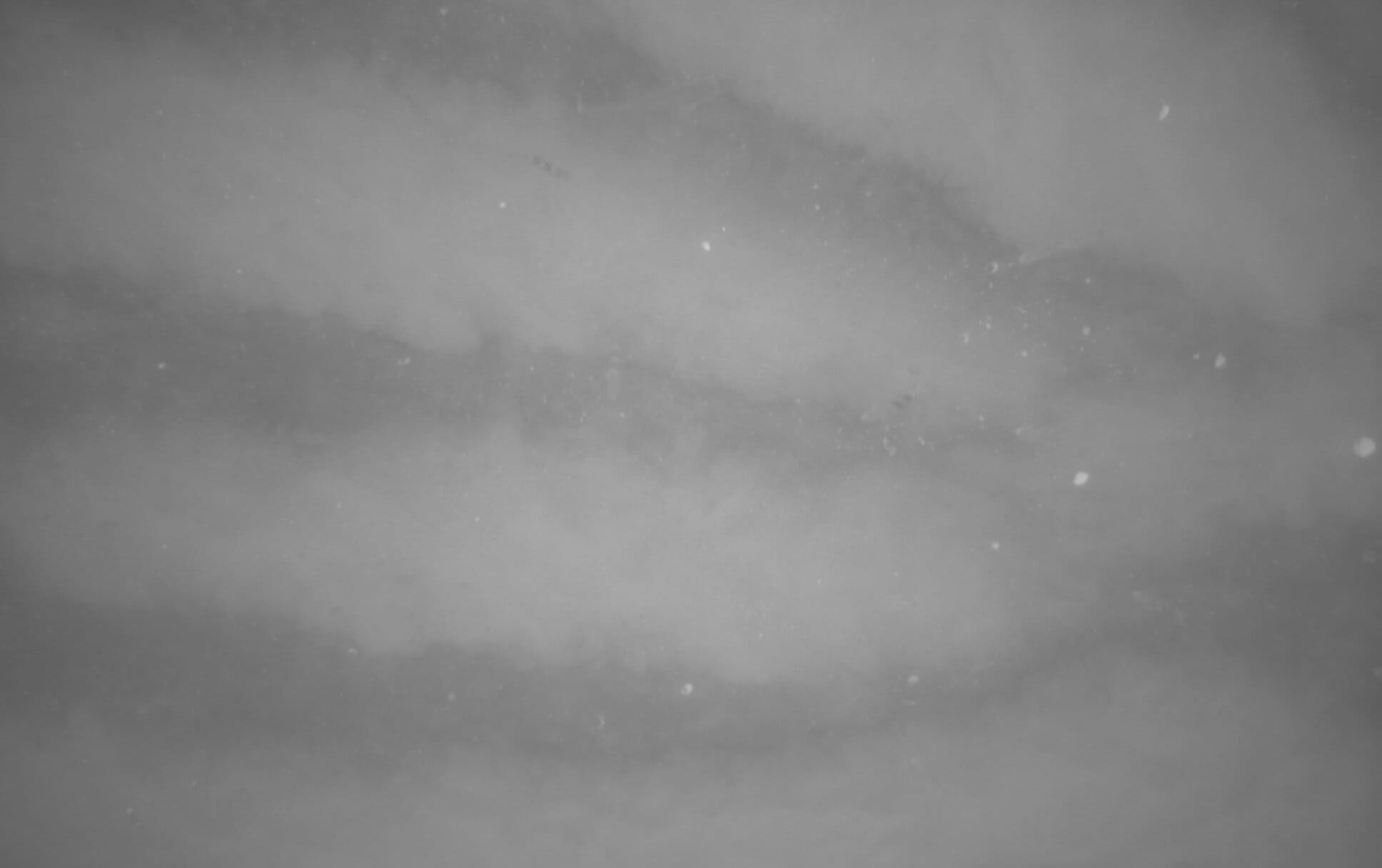}} 
        & \adjustbox{valign=m,vspace=.15mm}{\includegraphics[height=0.1\textwidth, width=0.2\textwidth]{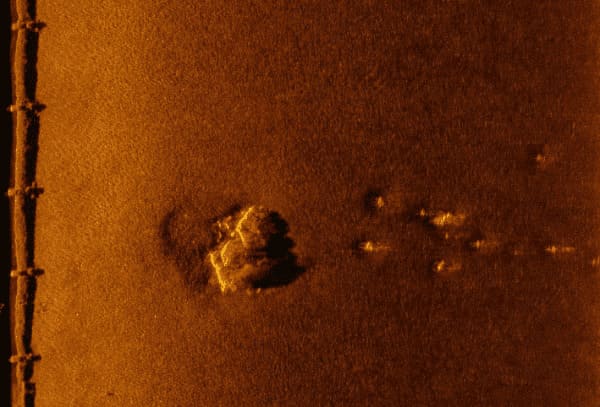}} \\
        \bottomrule
    \end{tabular}
    \caption{Overview of the data for each SubPipe chunk, including sample images from the cameras and the side-scan sonar.}
    \label{tab:subpipe-data}
\end{table*}

\begin{figure}[!t]
    \centering
    \includegraphics[width=.8\linewidth]{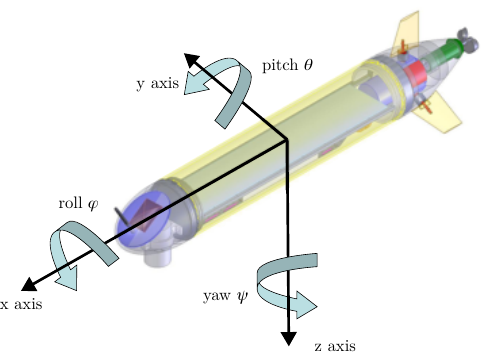}
    \caption{The 6 degrees of freedom of the LAUV \cite{vo:lsts_6dof}.}
    \label{fig:6DOF-lauv}
\end{figure}

SubPipe has been recorded during a survey mission performed by OceanScan-MST\footnote{OceanScan's home page: \url{https://www.oceanscan-mst.com/}}, near an underwater pipeline with about 1 km length in Porto, Portugal. The data recorded during the mission corresponds to the data measured by the onboard sensors and the estimated state of the robot as inferred from the sensors.

The onboard sensors are: two downward-looking cameras with 3-channel $2704 \times 1520$ and 1-channel $1936 \times 1216$ images, respectively; a Klein 3500 side-scan sonar, providing sonar images recorded at 900 KHz and 455 KHz, with a range of 30m per transducers, resulting in monochrome waterfall images with sizes $5000 \times 500$ and $2500\times 500$, respectively. Further details on the dataset, such as the exact number of frames and the intrinsic and extrinsic parameters of the cameras, the side-scan sonar, and other sensors, can be found in the online repository of SubPipe.

\begin{figure}[t]
  \centering

    \begin{tabular}{@{\hspace{0.3mm}}c@{\hspace{0.3mm}}c@{\hspace{0.3mm}}c@{\hspace{0.3mm}}c@{}}

      \adjustbox{valign=m,vspace=.05mm}{\includegraphics[width=0.12\textwidth]{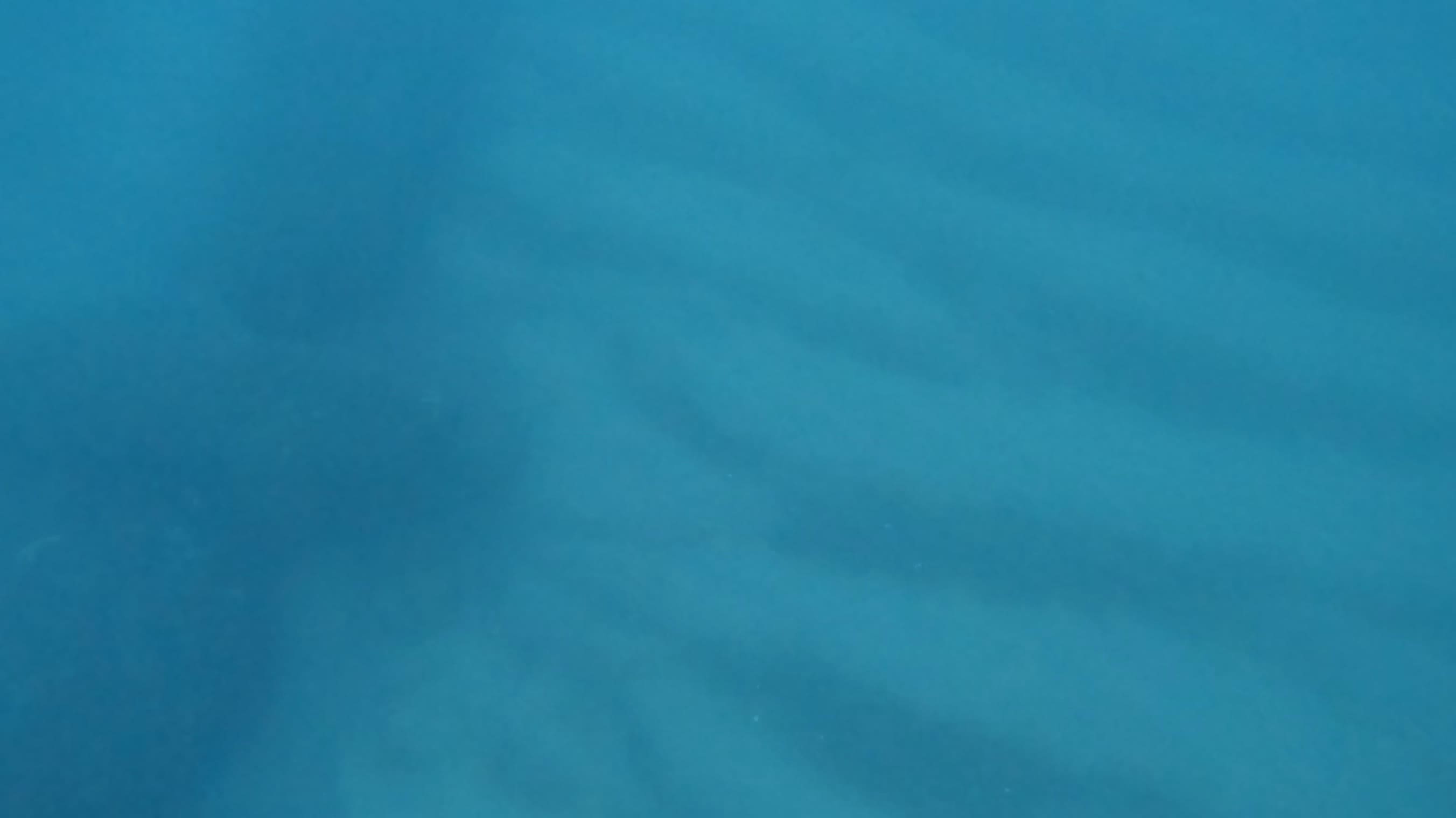}} &
      \adjustbox{valign=m,vspace=.05mm}{\includegraphics[width=0.12\textwidth]{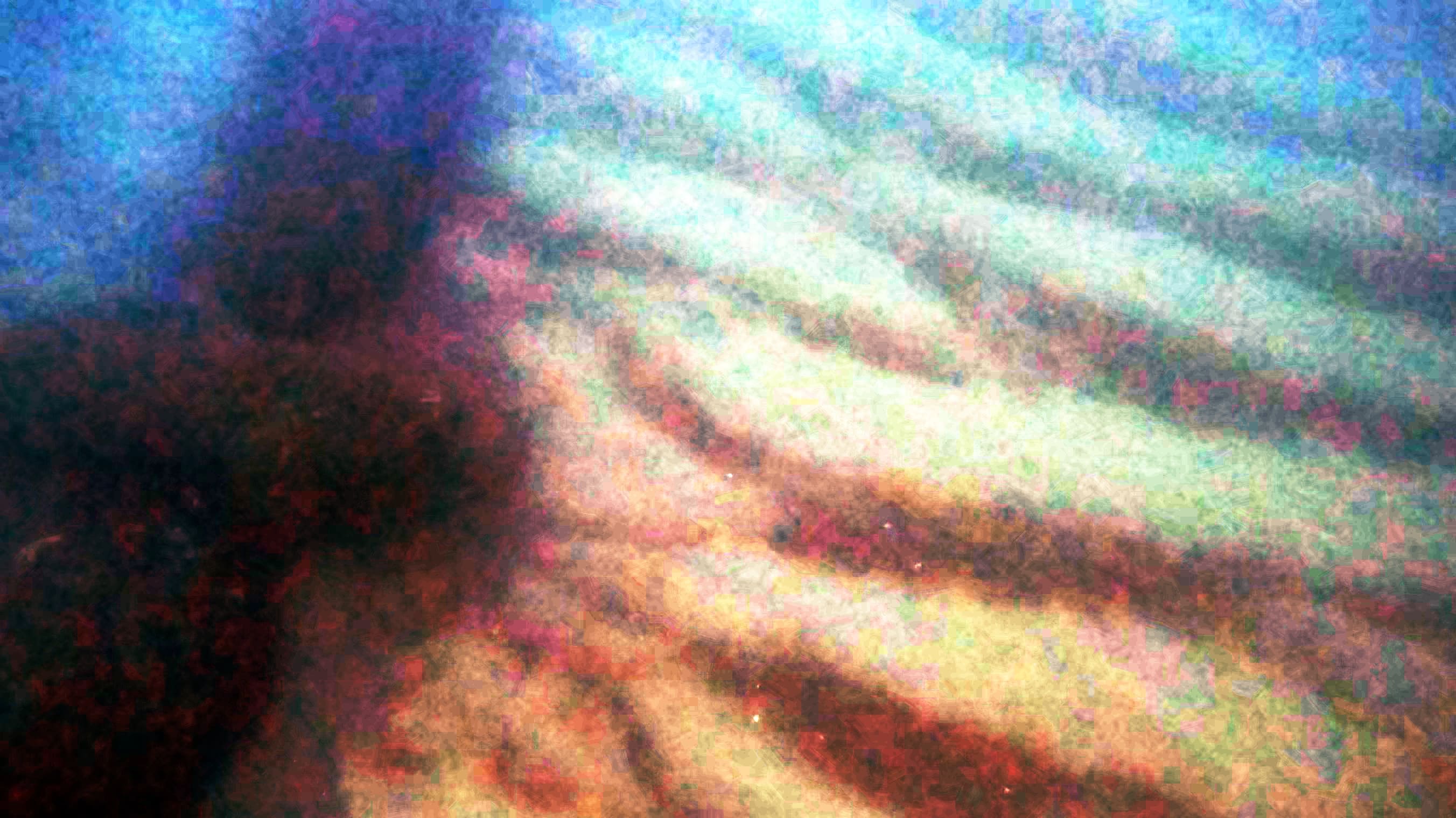}} &
      \adjustbox{valign=m,vspace=.05mm}{\includegraphics[width=0.12\textwidth]{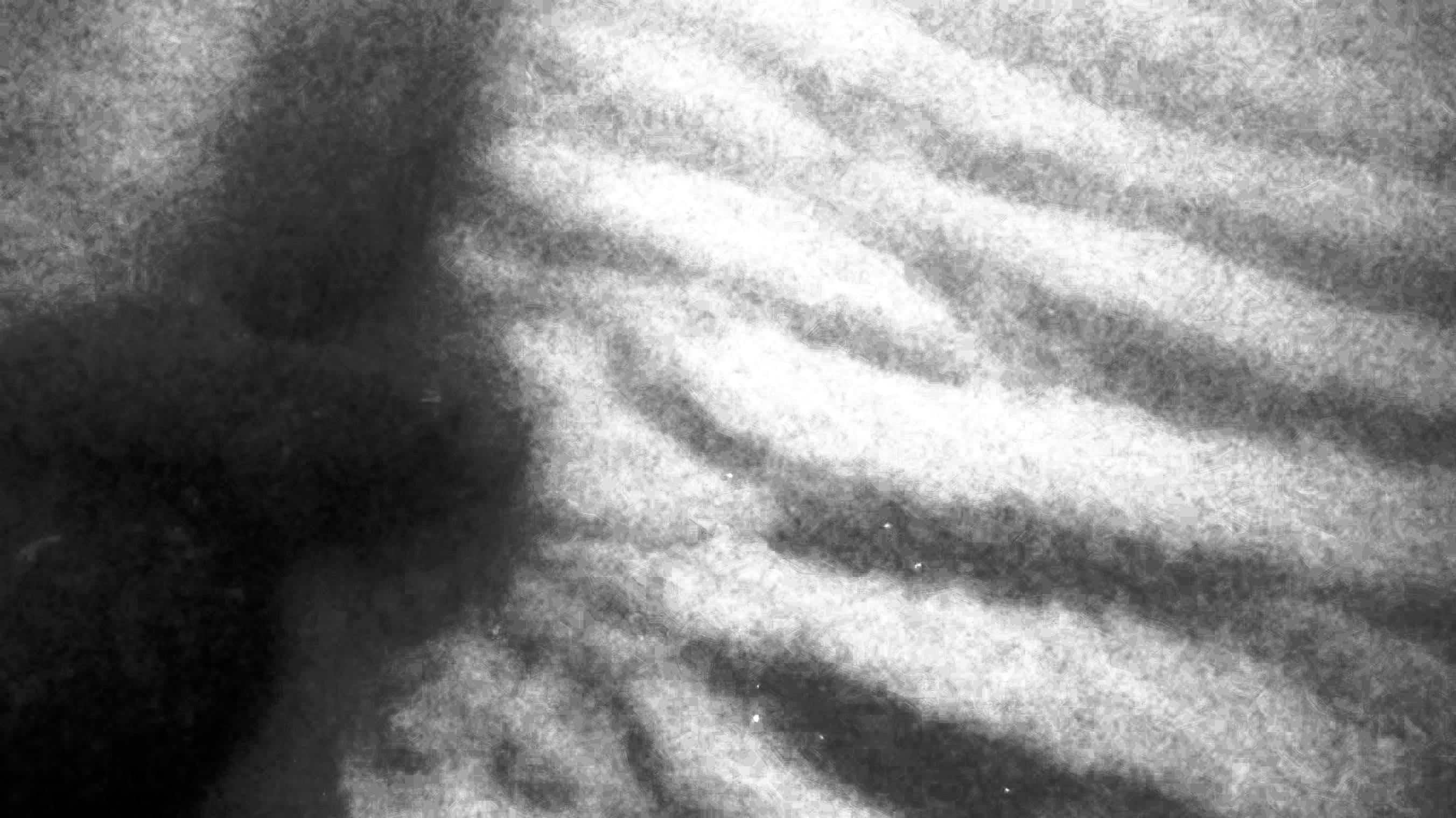}} &
      \adjustbox{valign=m,vspace=.05mm}{\includegraphics[width=0.12\textwidth]{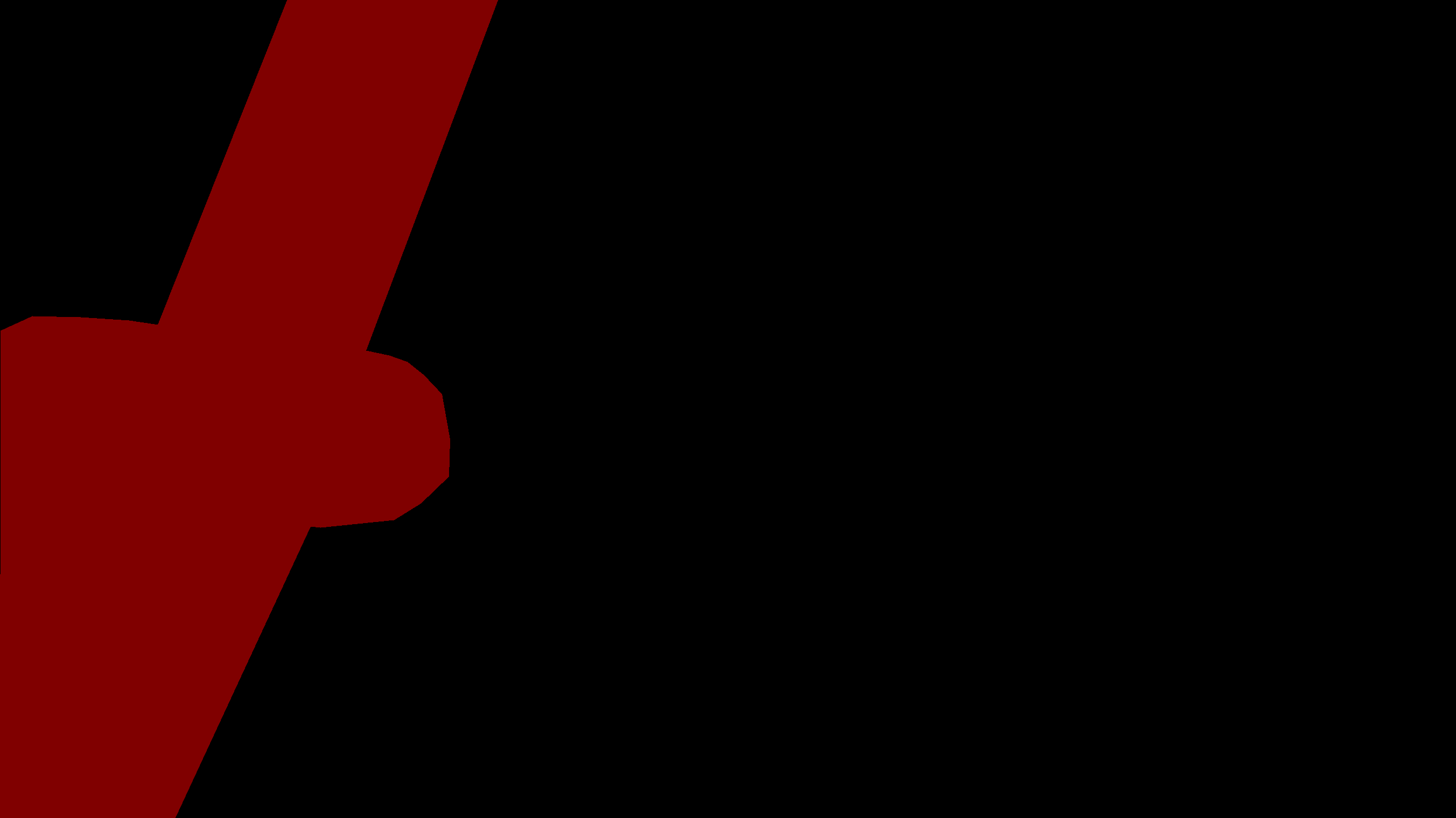}}
      \\[2.mm]

      \adjustbox{valign=m,vspace=.05mm}{\includegraphics[width=0.12\textwidth]{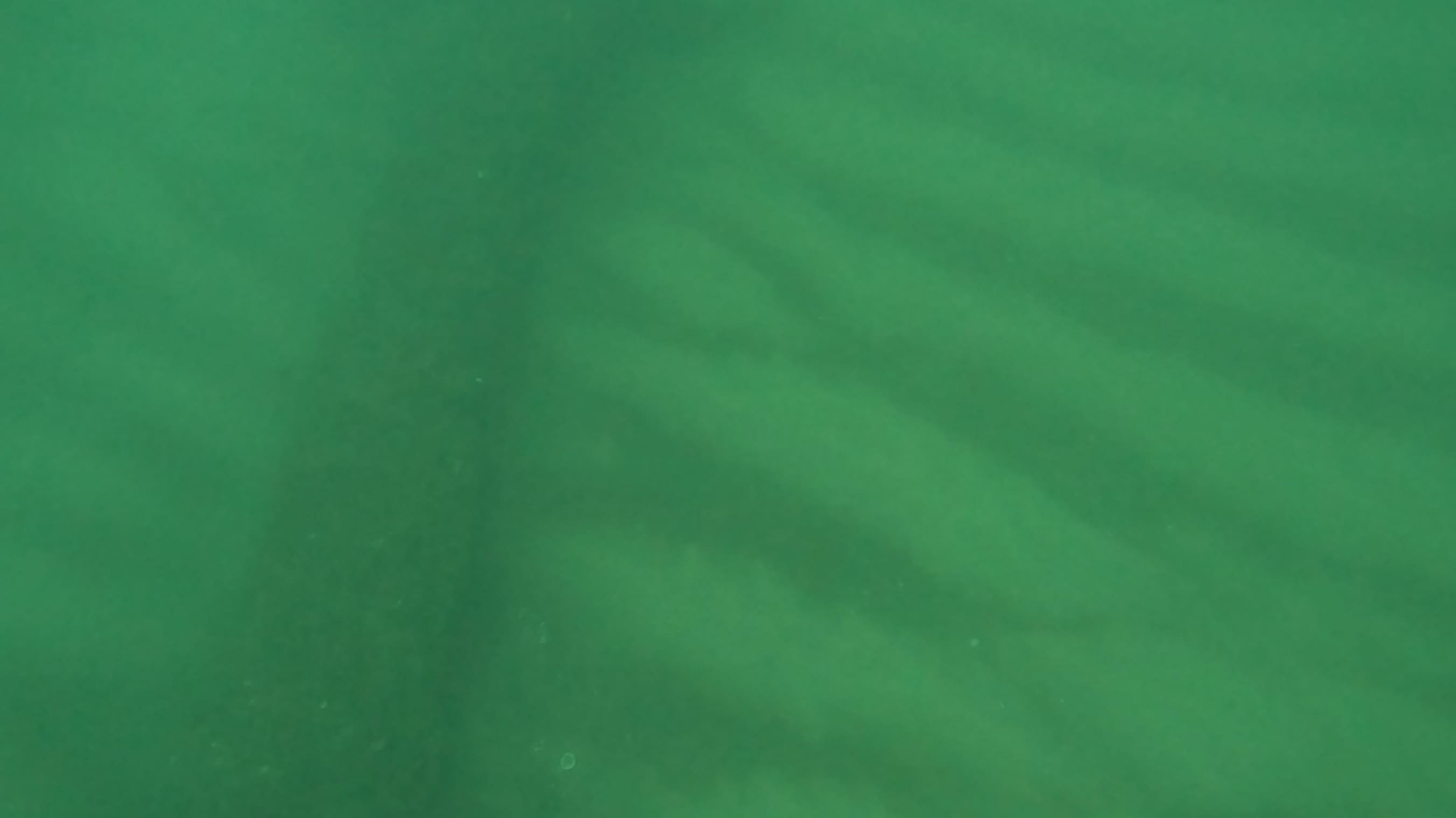}} &
      \adjustbox{valign=m,vspace=.05mm}{\includegraphics[width=0.12\textwidth]{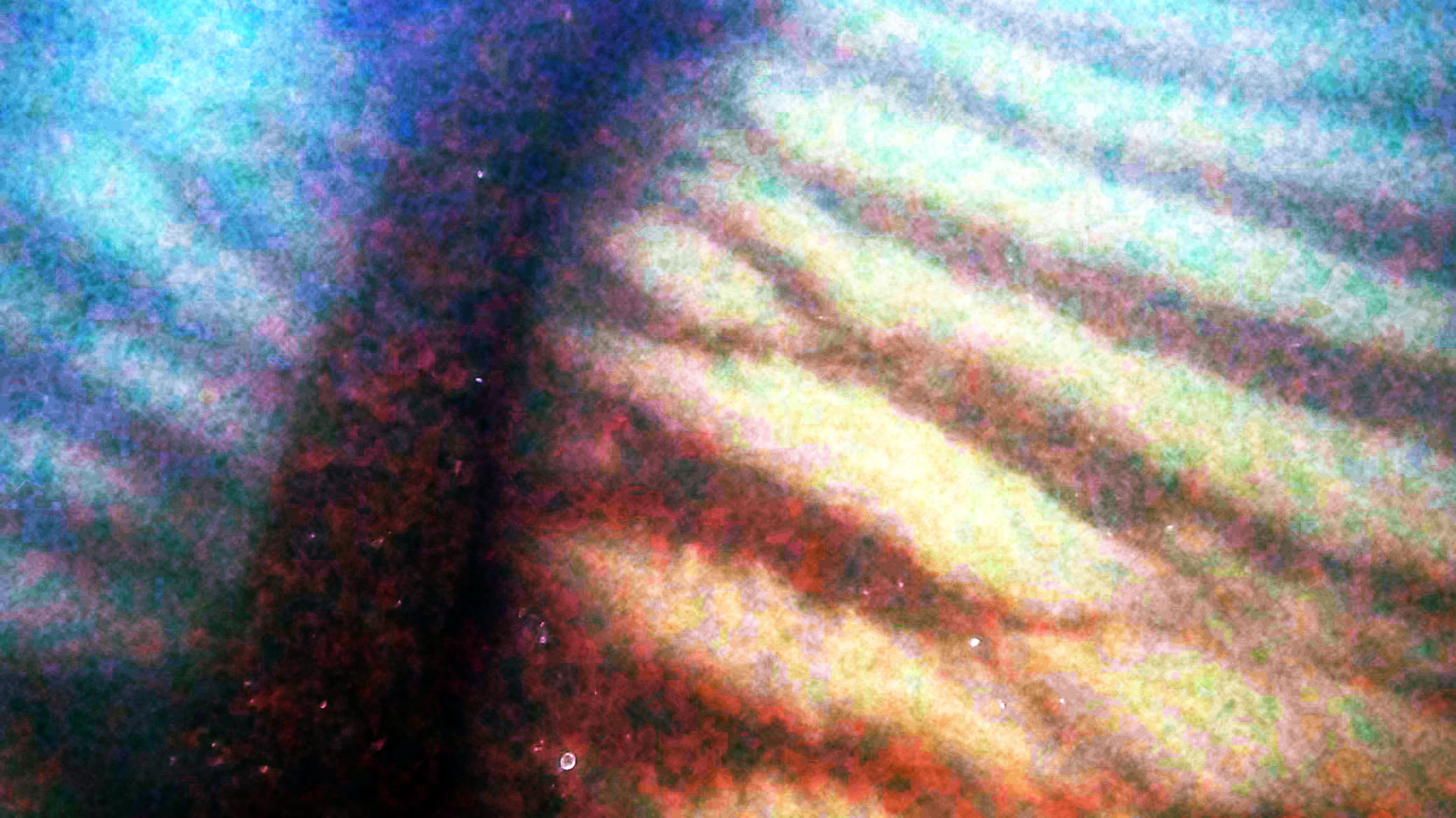}} &
      \adjustbox{valign=m,vspace=.05mm}{\includegraphics[width=0.12\textwidth]{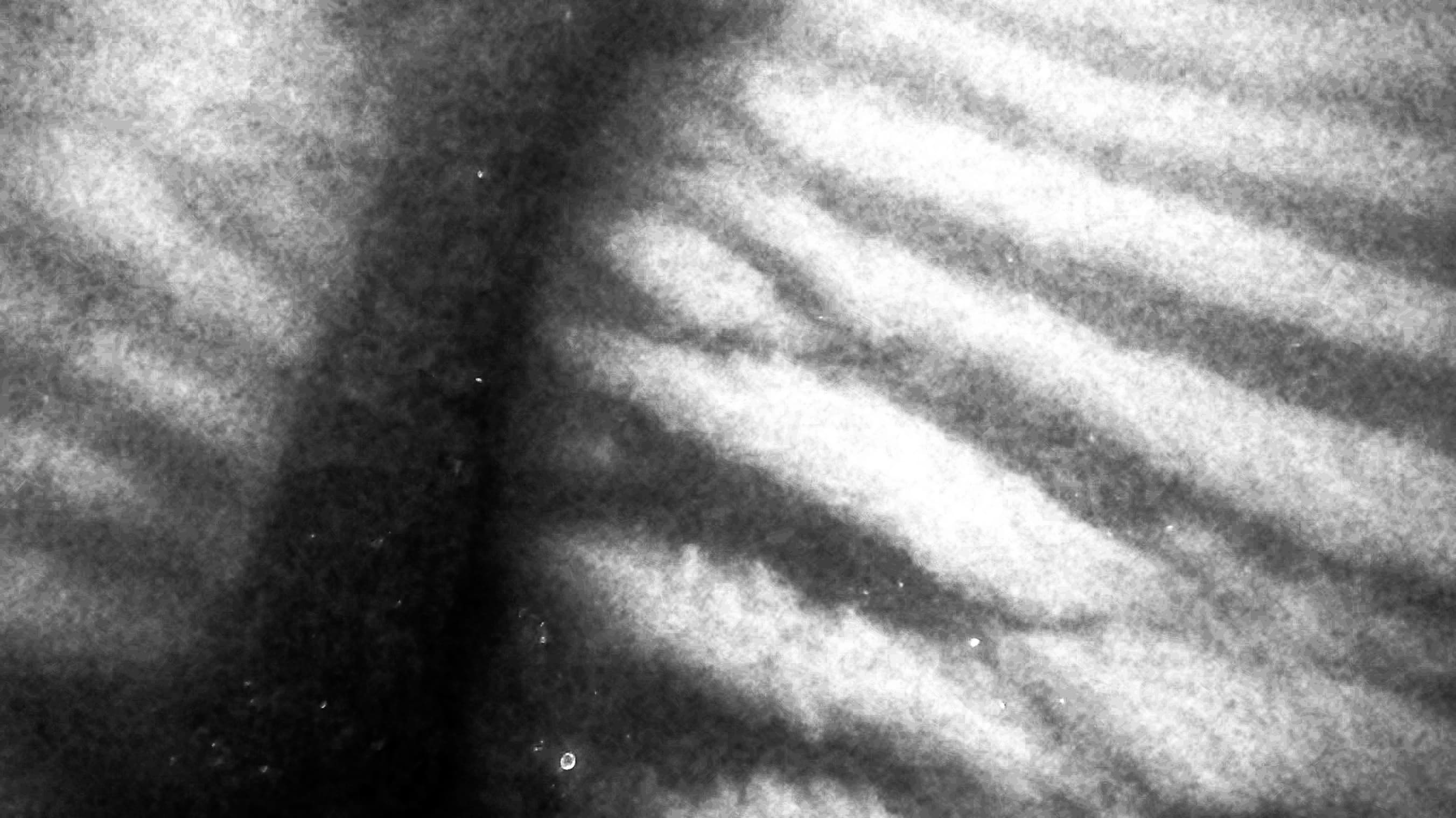}} &
      \adjustbox{valign=m,vspace=.05mm}{\includegraphics[width=0.12\textwidth]{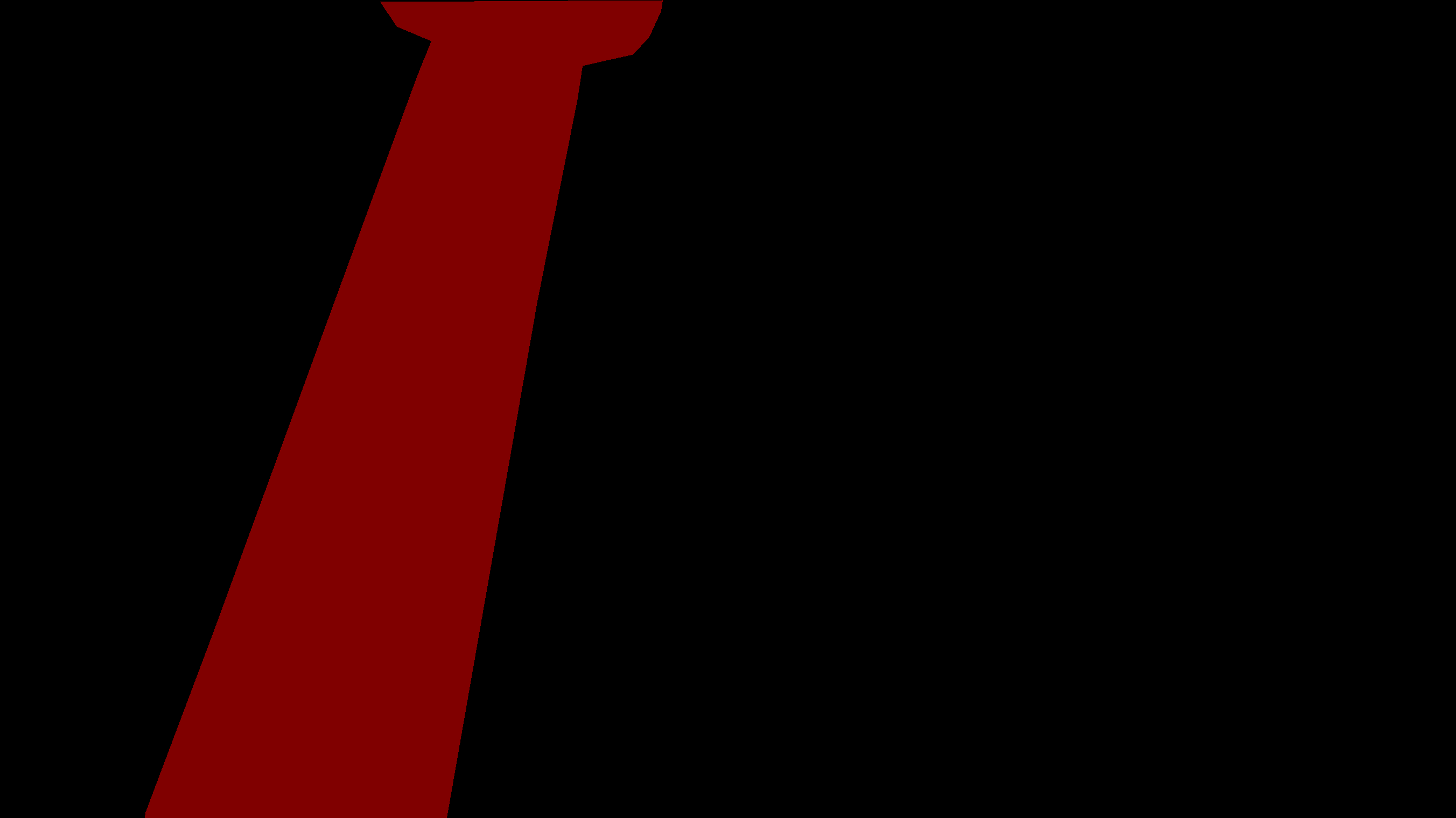}}
      \\[2.mm]

  \end{tabular}

  \caption{ Segmentation mask generation process illustrated with two examples. Each row presents a unique example. From left to right: the original raw image, the image after histogram equalization of RGB channels, the original image after converting to grayscale and applying histogram equalization, and the final segmentation mask. The equalization process enhances contrast, aiding in the precise delineation of the pipeline, including the pipe clamp, which is annotated as part of the pipeline.}
  \label{fig:segmentation-annotation}

\end{figure}

The estimated state corresponds to the 6 DOF pose of the robot (see \cref{fig:6DOF-lauv}) as inferred from the \gls{INS} and DVL. The three translational measurements, surge~$x$, sway~$y$, and heave~$z$, are recorded in meters and are estimated by a Kalman filter considering the last recorded GPS coordinates and measured velocities. On the other hand, the rotational measurements roll~$\phi$, pitch~$\theta$, and yaw~$\psi$ are provided in radians and measured by a gyroscope. Besides, SubPipe provides other measurements of i. the \gls{AUV} (e.g., depth, altitude, velocities, angular velocities, accelerations, rpm, forward distance measured from the nose of the \gls{AUV}) and ii. the environment (water velocity, temperature, pressure).


SubPipe provides five video sequences, ranging between approximately 7 and 9 minutes. The high-resolution camera has been recorded at 240 Hz, and the low-resolution one at 4 Hz. All the SSS waterfall images have been extracted and manually annotated using the COCO format for object detection. This process created a single class named \textit{Pipeline}. 
The sonar images were resized to a dimension of $640 \times 640$ pixels for compatibility with computer vision algorithms. 

An overview of the data contained in each of the sequences (or chunks) is depicted in  \cref{tab:subpipe-data}. Given the large dataset size, a smaller subsample of SubPipe referred to as SubPipeMini, is provided.
SubPipeMini contains approximately 20\% of the complete sequence data. Segmentation labels for the class 'pipeline' were manually annotated for every 25 frames of the 16170 high-resolution images of SubPipeMini, resulting in 647 labeled images. The annotation process was conducted using the LabelMe tool~\cite{Wada_Labelme_Image_Polygonal}. The repetitive nature of the underwater environment led to the decision to label the frames selectively, reducing redundancy in the labeled dataset.
The dataset's pipeline visibility is significantly reduced due to considerable blurring. To facilitate the labeling process, the images' contrast was enhanced using histogram equalization, as depicted in \cref{fig:segmentation-annotation}. Note that these preprocessed images, though crucial for label generation, are not part of the released dataset.

Since the high-resolution camera was not synchronized with the rest of the vehicle's sensors, no exact sensor measurements were available for the timestamps of the extracted video frames. To solve this issue, we used the recorded sensor measurements, and by applying linear interpolation, we calculated the estimated sensor values corresponding to the timestamp of each frame, formally:

\begin{equation}
    \nu^{*}(t) = \frac{\nu_1 - \nu_0}{t_1 - t_0} * (t - t_0) + \nu_0,
\end{equation}

\noindent where $\langle t_0, \nu_0 \rangle$ and $\langle t_1, \nu_1 \rangle$, represent the (closest in time) previous and following timestamp and measurement values, correspondingly. At the same time, $t$ is the timestamp for which we calculate the estimate, and finally, $\nu^{*}$ is the estimated value. This interpolation is applied for all sensor measurements, providing the estimated values for the timestamps corresponding to the extracted camera images.


\begin{figure*}[hbt]
\centering
\Huge
\resizebox{\textwidth}{!}{\begin{tabular}
{c c@{\hspace{.7mm}}c@{\hspace{3mm}}c@{\hspace{3mm}}c@{\hspace{3mm}}c@{\hspace{3mm}}c@{\hspace{3mm}}c@{\hspace{.3mm}}c@{\hspace{3mm}}c@{\hspace{.3mm}}c@{\hspace{3mm}}c@{\hspace{.3mm}}c}

 &\multicolumn{2}{c}{SubPipe} & \multicolumn{2}{c}{Aqualoc \cite{rw:dataset:aqualoc}} & \multicolumn{2}{c}{MIMIR \cite{rw:dataset:mimir}}
 &  \multicolumn{2}{c}{EuRoC \cite{rw:dataset:bernardi2022aurora}} & \multicolumn{2}{c}{KITTI \cite{dataset:kitti}}  & \multicolumn{2}{c}{TartanAir \cite{rw:dataset:tartanair2020iros}}  \\ 

\raisebox{.5cm}{\rotatebox[origin=l]{90}{delentropy (H)}} & \multicolumn{2}{c}{\includegraphics[width=.5\linewidth]{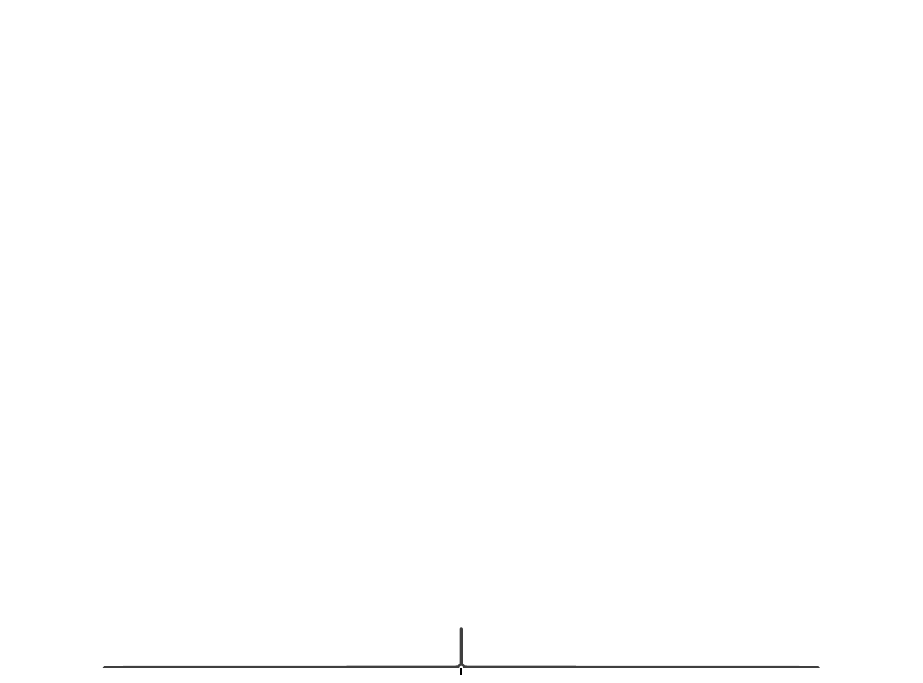}}
  & \multicolumn{2}{c}{\includegraphics[width=.5\linewidth]{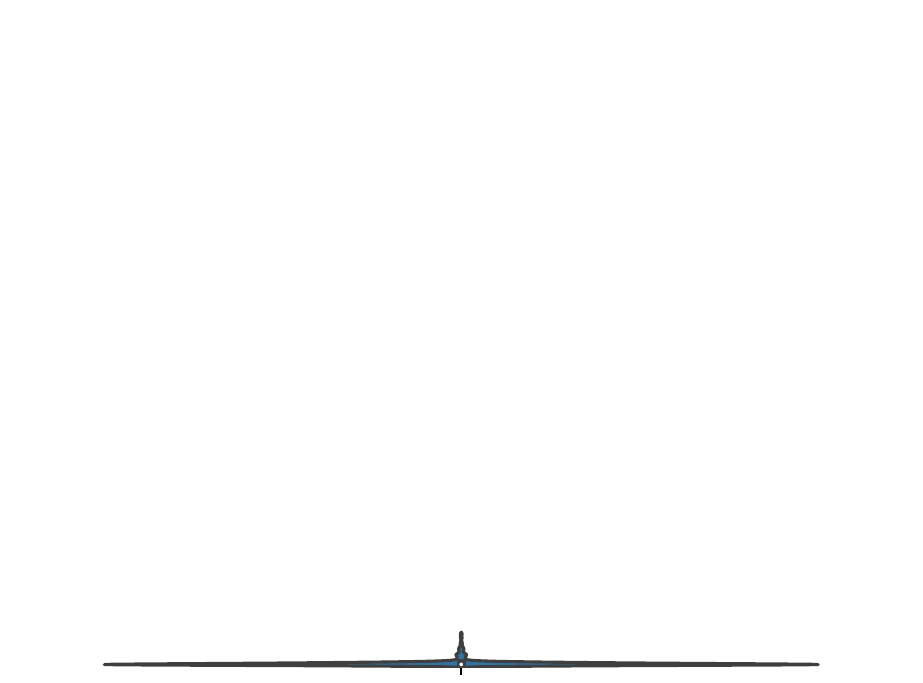}} 
  & \multicolumn{2}{c}{\includegraphics[width=.5\linewidth]{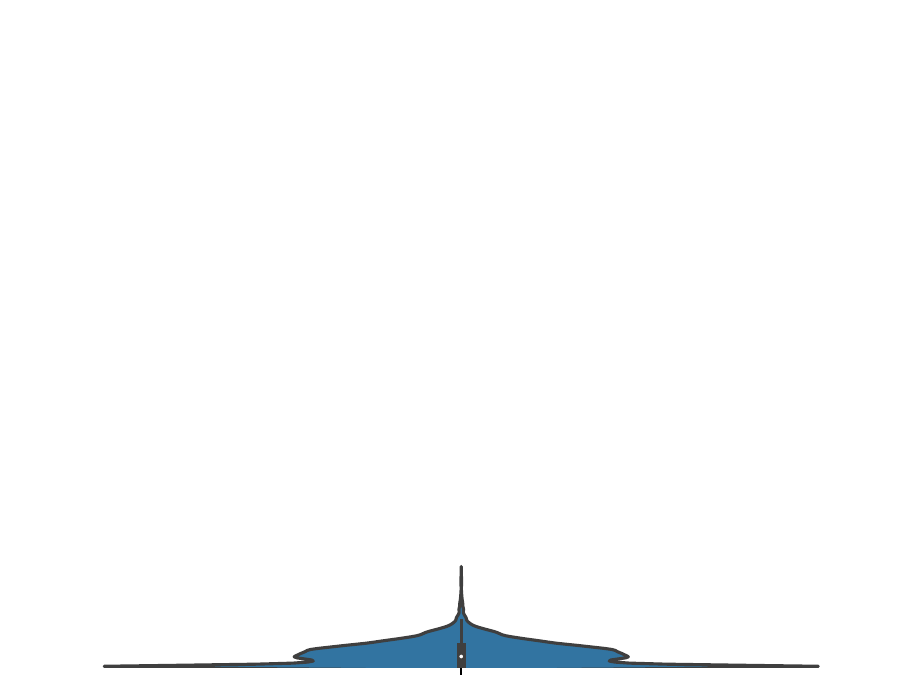}}
  &  \multicolumn{2}{c}{\includegraphics[width=.5\linewidth]{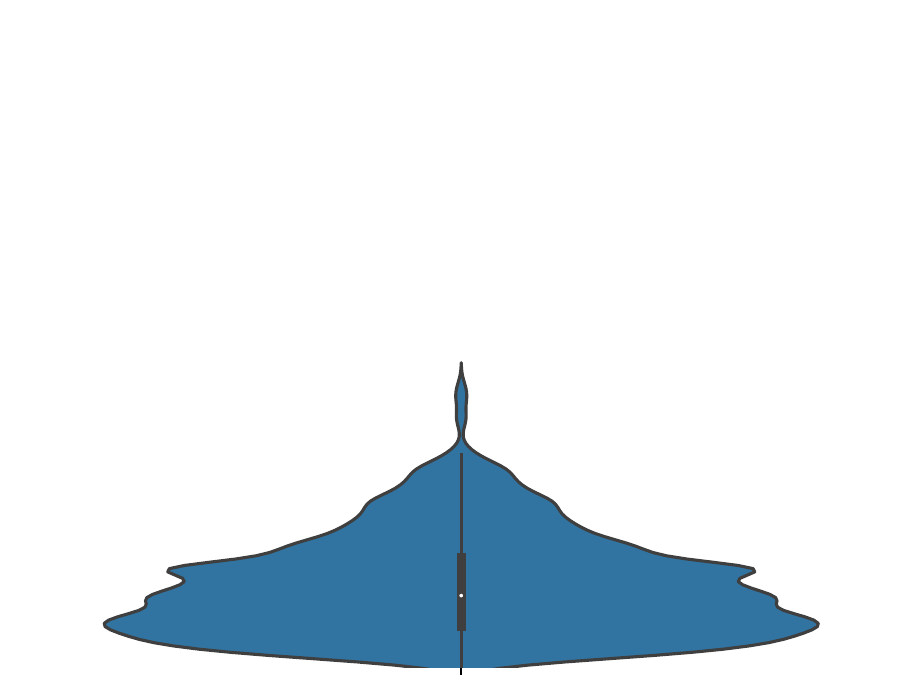}}  
  & \multicolumn{2}{c}{\includegraphics[width=.5\linewidth]{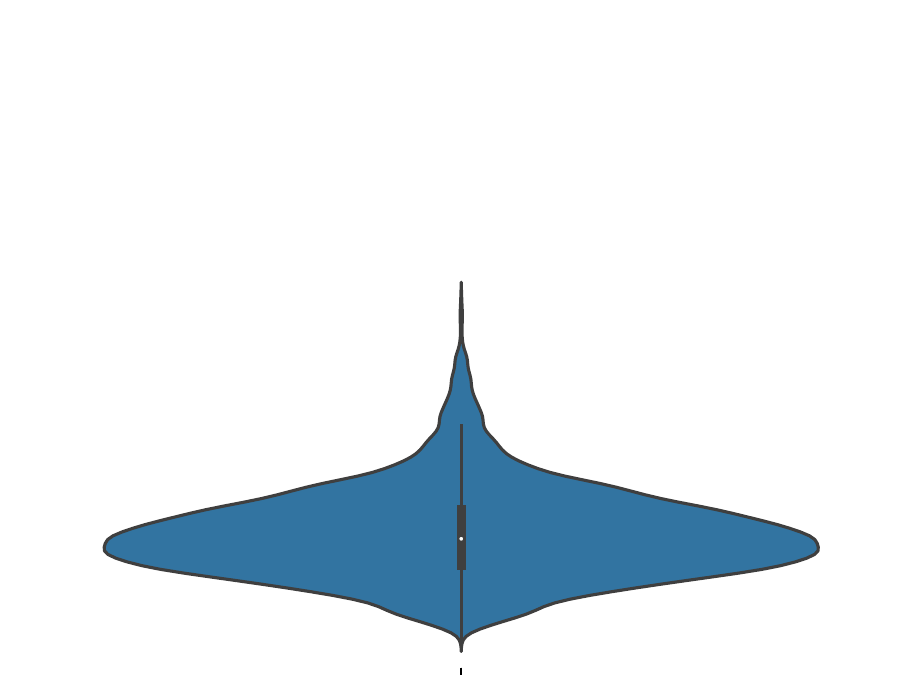}}    
  & \multicolumn{2}{c}{\includegraphics[width=.5\linewidth]{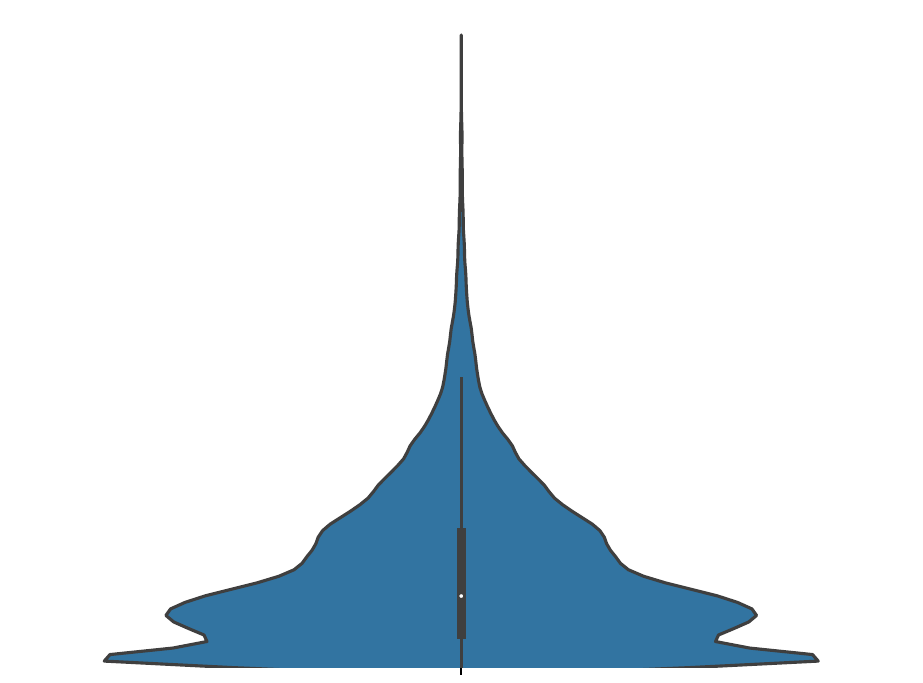}}   \\ 
 
 \rotatebox[origin=c]{90}{min/max H} &\adjustbox{valign=m,vspace=.2pt}{\includegraphics[height=.2\linewidth]{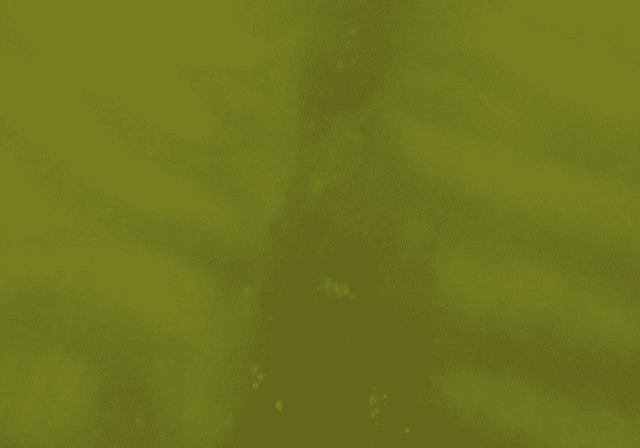}} 
& \adjustbox{valign=m,vspace=.2pt}{\includegraphics[height=.2\linewidth]{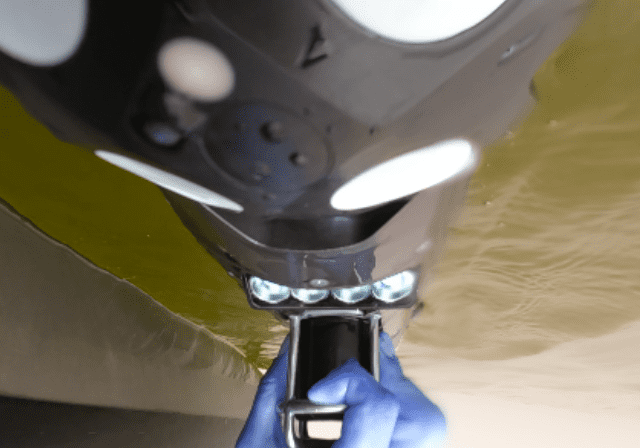}}

& \adjustbox{valign=m,vspace=.2pt}{\includegraphics[height=.2\linewidth]{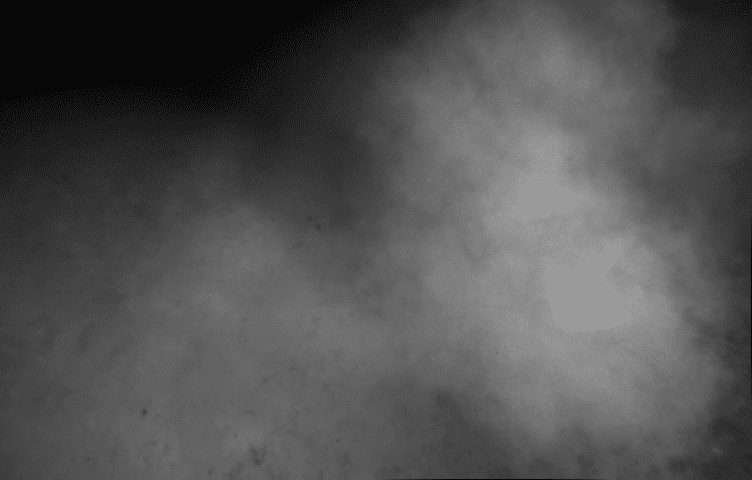}}
& \adjustbox{valign=m,vspace=.2pt}{\includegraphics[height=.2\linewidth]{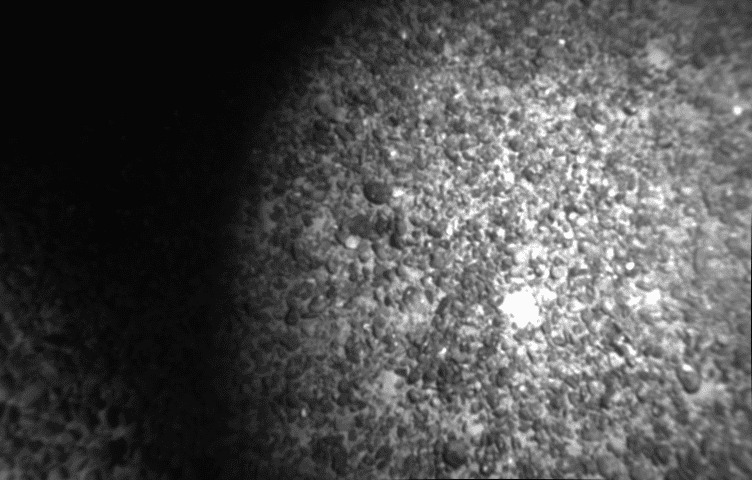}}

& \adjustbox{valign=m,vspace=.2pt}{\includegraphics[height=.2\linewidth]{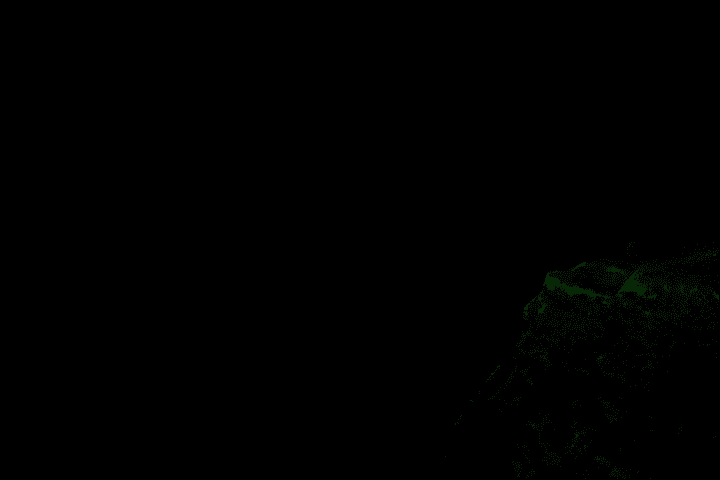}}
& \adjustbox{valign=m,vspace=.2pt}{\includegraphics[height=.2\linewidth]{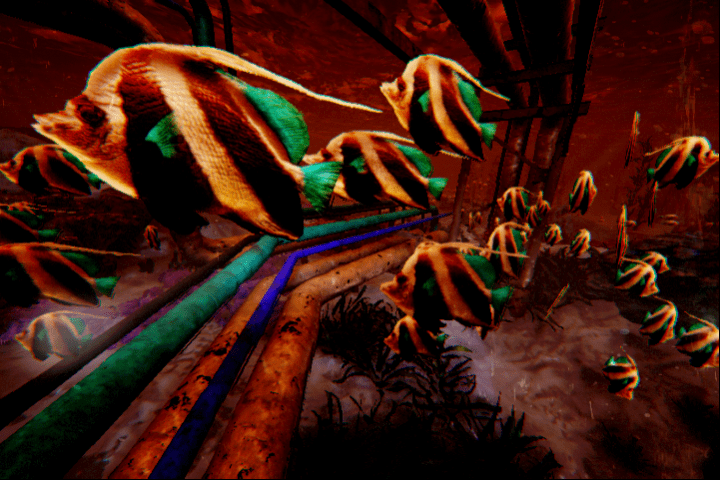}}

& \adjustbox{valign=m,vspace=.2pt}{\includegraphics[height=.2\linewidth]{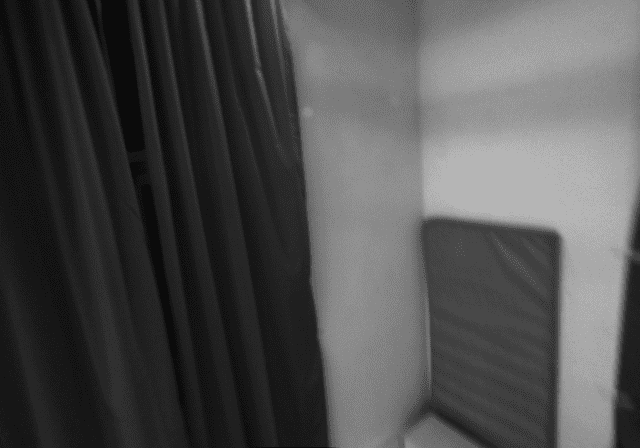}}
& \adjustbox{valign=m,vspace=.2pt}{\includegraphics[height=.2\linewidth]{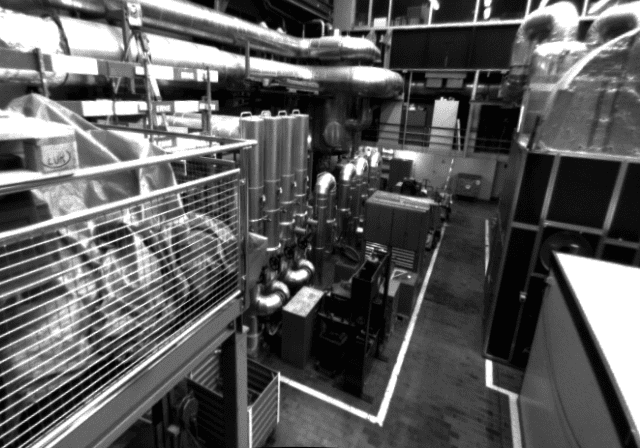}}

& \adjustbox{valign=m,vspace=.2pt}{\includegraphics[height=.2\linewidth, width = .4\linewidth]{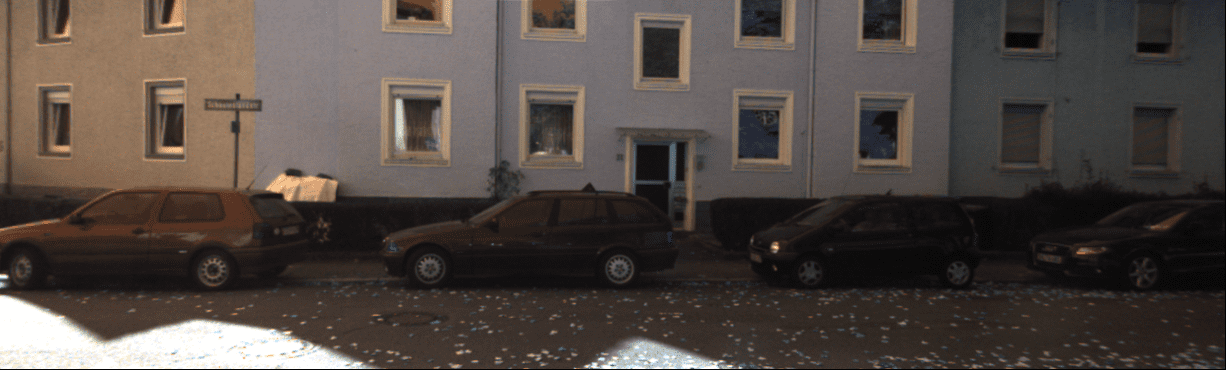}}
& \adjustbox{valign=m,vspace=.2pt}{\includegraphics[height=.2\linewidth, width = .4\linewidth]{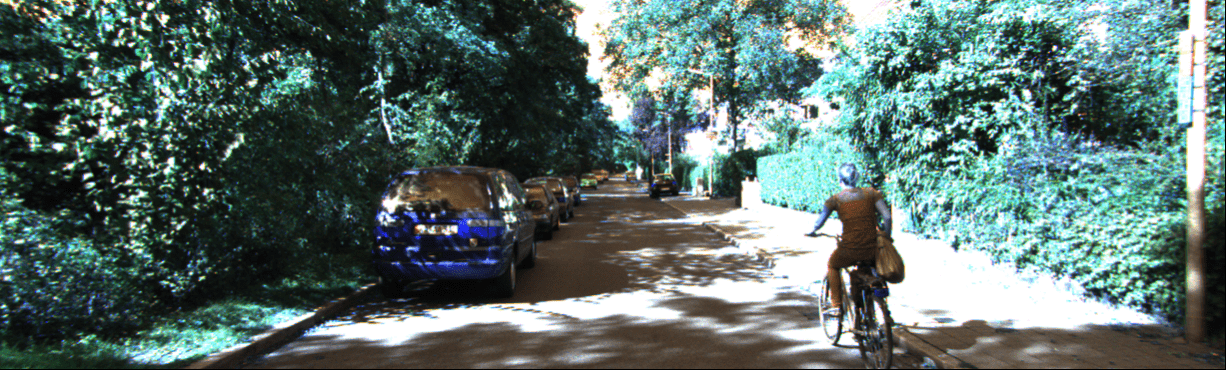}}

& \adjustbox{valign=m,vspace=.2pt}{\includegraphics[height=.2\linewidth]{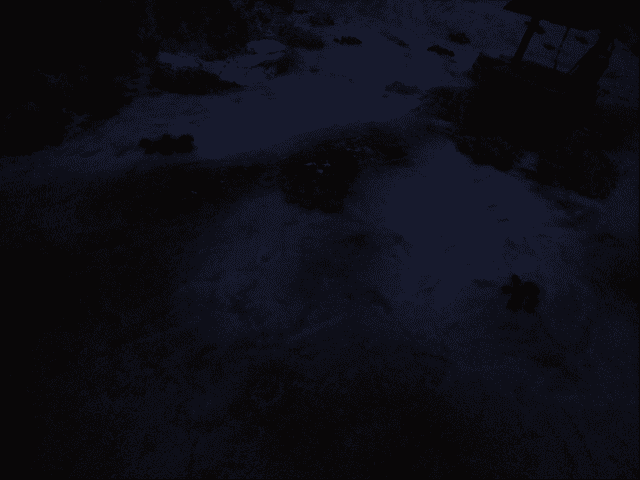}} 
& \adjustbox{valign=m,vspace=.2pt}{\includegraphics[height=.2\linewidth]{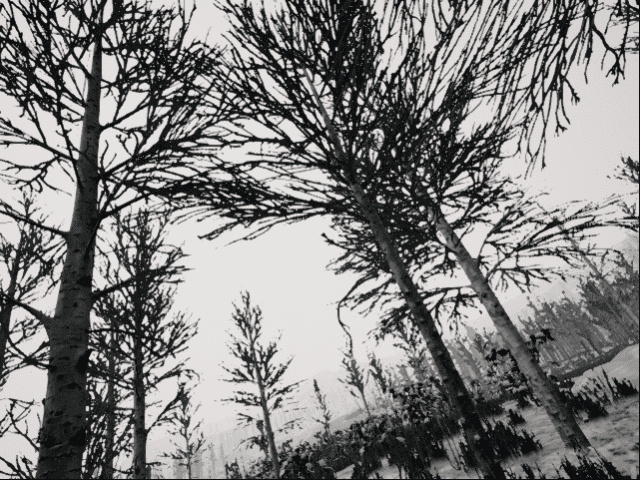}} \\

\rotatebox[origin=c]{90}{$\sigma$} &\multicolumn{2}{c}{0.05} & \multicolumn{2}{c}{0.14} & \multicolumn{2}{c}{0.35}
 &  \multicolumn{2}{c}{0.44} & \multicolumn{2}{c}{0.014}  & \multicolumn{2}{c}{0.49}  \\ 
 
\end{tabular}}
\caption{Dataset metrics. The top section presents the distribution of delentropy values across various datasets, accompanied by representative images that yield the minimum and maximum delentropy. On the bottom are the motion diversity metric results. Lower delentropy and motion diversity values indicate a high degree of uniformity in image content and limited motion variety across the six degrees of freedom. This phenomenon is particularly evident in SubPipe, since pipeline inspection missions inherently limit motion and imaging variability. In contrast, datasets such as EuRoC, featuring a drone navigating freely in six degrees of freedom within an indoor environment, exhibit significantly more diversity in both imaging and motion.}
\label{fig:datasetmetrics}
\end{figure*}

\section{Dataset metrics}
\label{section:metrics}
The information contained by the dataset is quantitatively evaluated and compared with existing state-of-the-art datasets using the metrics proposed in \cite{rw:dataset:mimir} and \cite{rw:dataset:tartanair2020iros}: delentropy and motion diversity.

The delentropy (or image entropy) $H$ is calculated for each image in the dataset to measure the amount of information it contains. It is based on Shannon's joint entropy formulation, taking the image's gradient in the $x$ and $y$ axes as the joint probability density function. A uniform distribution of the joint probability density function for the gradient yields maximum values for $H$. The delentropy is maximized under the presence of nondistinctive features and minimized (zeroed) for zero-gradient images containing no features. Ideal conditions are then represented by intermediate values that indicate the presence of distinctive features.

The motion diversity metric $\sigma$ is a function of the principal components of the sequence of relative motions. The motion diversity metric converges to one with evenly distributed motion sequences, and to zero under the presence of motion in only one axis.

Figure \ref{fig:datasetmetrics} depicts the result of deploying the proposed metrics on SubPipe and the datasets for visual localization Aqualoc \cite{rw:dataset:aqualoc}, MIMIR \cite{rw:dataset:mimir}, EuRoC \cite{dataset:burri2016euroc}, KITTI \cite{dataset:kitti}, and TartanAir \cite{rw:dataset:tartanair2020iros}.  This deployment is driven by the specific design of these metrics for localization datasets. Aqualoc and MIMIR are real and synthetic underwater datasets, respectively. TartanAir includes a high diversity of synthetic tracks, including underwater environments. EuRoC and KITTI are above-water datasets recorded with a drone and a car, respectively.
Noticeably, above-water datasets retrieve much more information-rich images than underwater ones. Underwater datasets have a higher presence of images with low sharpness due to textureless or blurred areas caused by the seabed's uniformity and light scattering, among other phenomena. Nevertheless, motion blur and low illumination frames can also retrieve low entropy values above water, as seen in EuRoC. MIMIR presents richer images among the underwater datasets, as is to be expected in simulated images. SubPipe's images present delentropy values very close to zero in almost all images, originating from the uniformity of the background and the lowly illuminated sequences.
Regarding motion diversity, drone sequences like the ones in EuRoC and TartanAir present higher diversity. SubPipe, however, presents a low diversity, as is expected for a pipeline inspection mission: the robot follows a mainly straight path, only altered by the seabed's and the pipe's shape, which the robot must follow. In that sense, the diversity of the motion is similar to a car's motion, with the motion, in this case, constrained by the degrees of freedom of the car. That is evidenced by the similarity between SubPipe and KITTI's motion diversity metrics.

\section{Experimental evaluation}
\label{sec:evaluations}

This section proposes a set of experimental evaluations for SubPipe, consisting of leveraging state-of-the-art algorithms for RGB image segmentation, visual SLAM, and object detection in side-scan sonar images.

\subsection{Visual SLAM}
The proposed state-of-the-art SLAM algorithms are the geometry-based algorithms ORB-SLAM3 \cite{experiment:orbslam3} and DSO~\cite{experiment:dso}, and the learning-based method TartanVO \cite{vo:wang2021tartanvo}. 

ORB-SLAM3 integrates an indirect-based approach that relies on ORB descriptors that are efficiently tracked across frames following a constant velocity model. Meanwhile, an Atlas map is created, in which, under tracking loss, the current map is stored, and a new map is created. These maps are merged into one if a loop is detected.  Indirect methods like ORB-SLAM3 rely on descriptor matching and thus are prone to fail in low-textured areas such as the ones present in SubPipe. Conversely, direct methods like the proposed DSO compare pixel intensities across image frames, therefore using more information from the image. These methods are more robust to low-textured areas but more sensitive to large baselines. DSO, in particular, presents a sparse approach that tracks all those pixels in the image with a high gradient.
The learning-based method TartanVO, as described in \cite{vo:wang2021tartanvo}, leverages PWC-Net \cite{sun2018pwcnet} for computing optical flow and employs a tailored version of ResNet50 \cite{dl:vo:2015resnet} for pose estimation, featuring dual output branches dedicated to rotation and translation respectively. Furthermore, it integrates the intrinsic camera parameters via an intermediary layer that bridges the optical flow computation and pose estimation components, enhancing its adaptability to various camera parameters. The loss function proposed is a combination of optical flow and camera motion.

\begin{figure}
    \centering
    \includegraphics[width=1\linewidth]{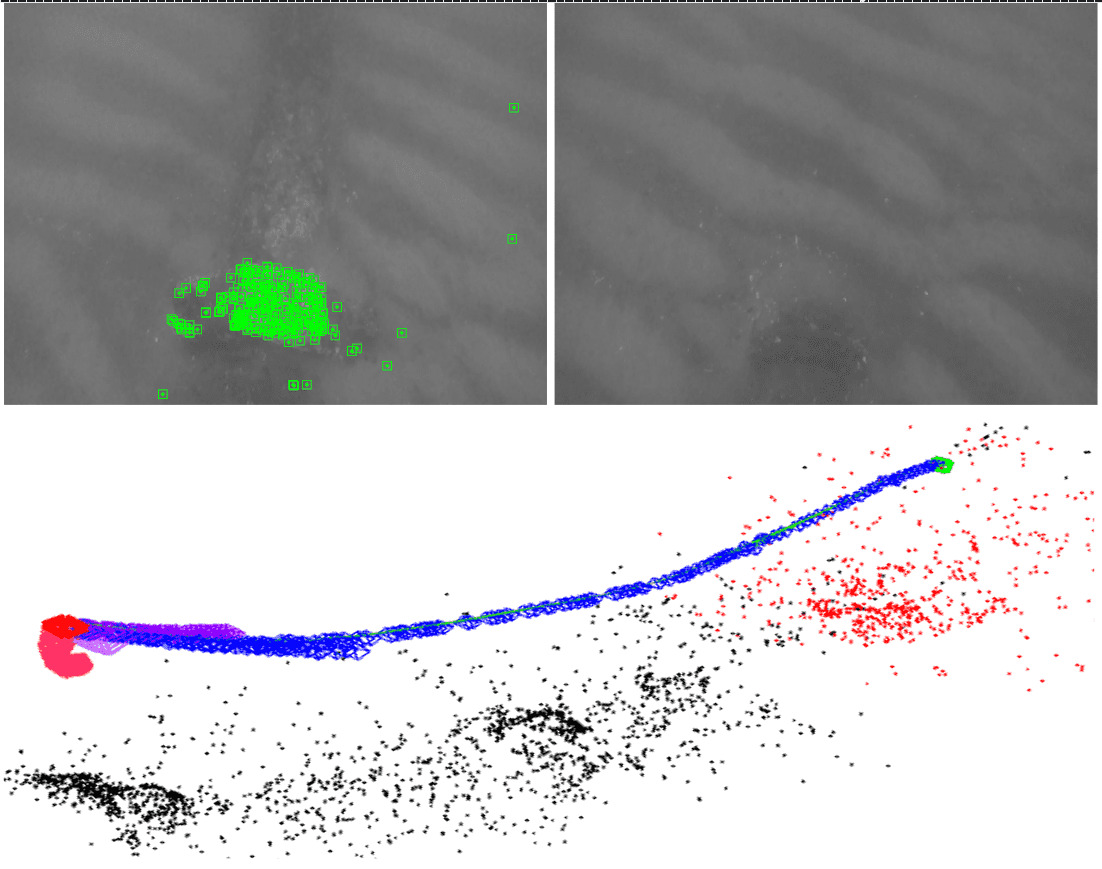}
    \caption{ORB-SLAM3's performance on SubPipe. The pipe clamps (top-left) provide enough features to track the camera's motion; however, the track is lost once the clamps and the pipe are out of sight (top-right).}
    \label{fig:orbslam3}
\end{figure}

\begin{figure}
    \centering
    \includegraphics[width=1\linewidth]{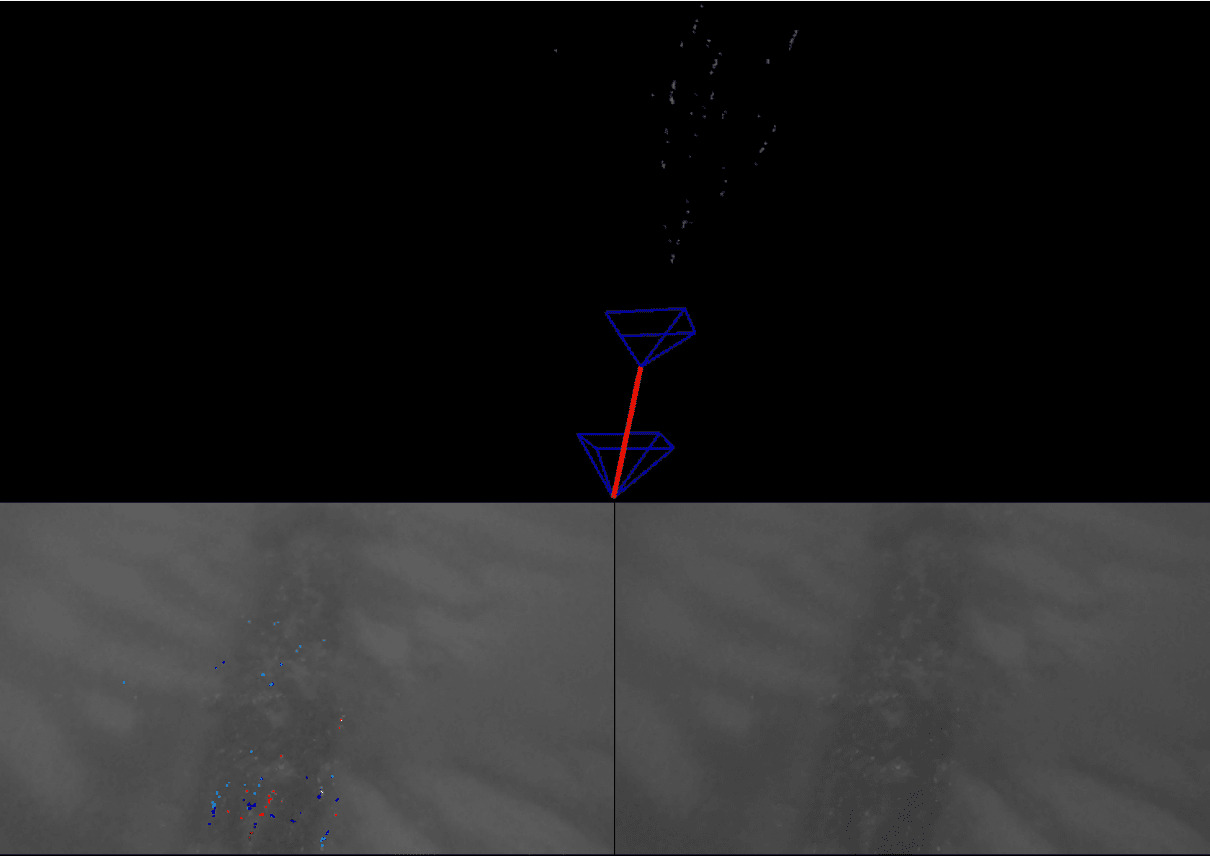}
    \caption{DSO's performance on SubPipe. The low gradients do not allow the detection of enough features (the colored pixels in the bottom-left image) to be tracked after the algorithm's initialization.}
    \label{fig:dso}
\end{figure}

Both geometry-based algorithms fail to track the trajectory under SubPipe's very challenging imaging conditions. The lack of features and gradients in the image, as evidenced by the delentropy metric in Section \ref{section:metrics}, is the source of failure. ORB-SLAM3 can track some pipe sections, as depicted in Fig. \ref{fig:orbslam3}. The pipe clamps provide a higher density of distinctive features that the algorithm can track. However, the track is lost as soon as those areas are out of the cameras' field of view. Similarly, the uniformity of the pixel gradients does not provide DSO with enough points to track across frames, as shown in Fig. \ref{fig:dso}.

\begin{figure}
    \centering
    \includegraphics[width=1\linewidth]{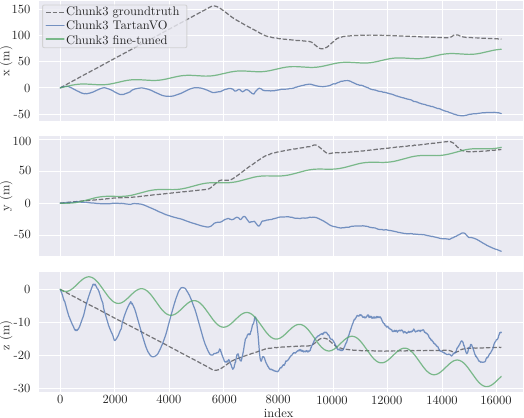}
    \caption{Results of applying TartanVO's models in SubPipe's Chunk 3. In blue, the original model provided by the authors. In green, the original model fine-tuned with SubPipe's Chunk 2. }
    \label{fig:results-tartanvo-chunk3}
\end{figure}

The lack of information on SubPipe's images challenges the performance of geometry-based algorithms. Under these conditions, there is a potential for visual localization algorithms to benefit from the higher-level representations of the image that learning-based approaches can achieve \cite{olayasurvey}. Thus, the next set of proposed experiments takes TartanVO as the baseline architecture for leveraging learning-based visual odometry architectures.
The experiments involve leveraging TartanVO's original model on SubPipe first and then a fine-tuned model. 
The last layers of the optical flow module and the pose regressor were fine-tuned with SubPipe's chunk 2, which was afterward tested in chunk 3. Given that SubPipe does not provide optical flow ground truth, the loss function corresponds to the chordal loss for the SE(3) pose introduced in \cite{olaya2023loss}. It was fine-tuned for $15$ epochs and with an Adam optimizer with an initial learning rate of $10^{-4}$ and weight decay factor of $10^{-3}$.
Table \ref{table:results-tartanvo-chunk3} shows the \gls{APE} and the \gls{RPE} \cite{metrics:sturm2012benchmark} metrics resulting from comparing the ground-truth with the results inferred from both models. While the error is considerably lower in the fine-tuned model than in the original model, Fig. \ref{fig:results-tartanvo-chunk3} evidences that such low error indicates that the model overfitted to the data. This is caused by the uniformity of the images and the trajectory, as evidenced by the dataset metrics in Section \ref{fig:datasetmetrics}. Therefore, additional datasets in the fine-tuning process must be included, introducing a broader data distribution.

\begin{table}[ht!]
\centering
\footnotesize
\caption{Visual Odometry results on SubPipe's chunk 3.}
\label{table:results-tartanvo-chunk3}
\begin{tabular}{cc cc}
\toprule
\multicolumn{2}{c}{TartanVO (original model)} & \multicolumn{2}{c}{TartanVO (fine tuned w. SubPipe)}\\
\cmidrule(l{1em}r{1em}){1-2}\cmidrule(l{1em}r{1em}){3-4}
ATE[m] & RPE[m]  &  ATE[m] & RPE[m] \\
\midrule
182.3 & 0.4 & 66.5 & 0.02 \\

\bottomrule
\end{tabular}
\end{table}

\begin{table}
\caption{Sample results (Top) and mIoU (Bottom) for segmentation.}
\label{fig:segmentationresults}
\centering
\footnotesize
\resizebox{\linewidth}{!}{\begin{tabular}
{c rcc}
\toprule
Image & \multicolumn{1}{c}{Groundtruth}& SegFormer~\cite{xie2021segformer} &  DeepLabV3~\cite{chen2017rethinking} \\ 

 \adjustbox{valign=m,vspace=.2pt}{\includegraphics[width=.2\linewidth]{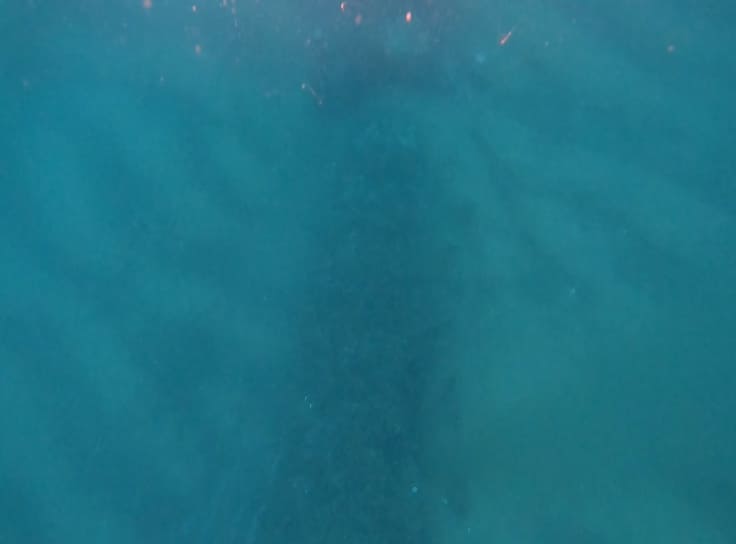}}
 
 & \adjustbox{valign=m,vspace=.2pt}{\includegraphics[width=.2\linewidth]{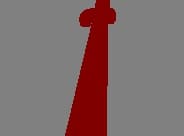}}

 & \adjustbox{valign=m,vspace=.2pt}{\includegraphics[width=.2\linewidth]{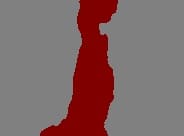}} 

 & \adjustbox{valign=m,vspace=.2pt}{\includegraphics[width=.2\linewidth]{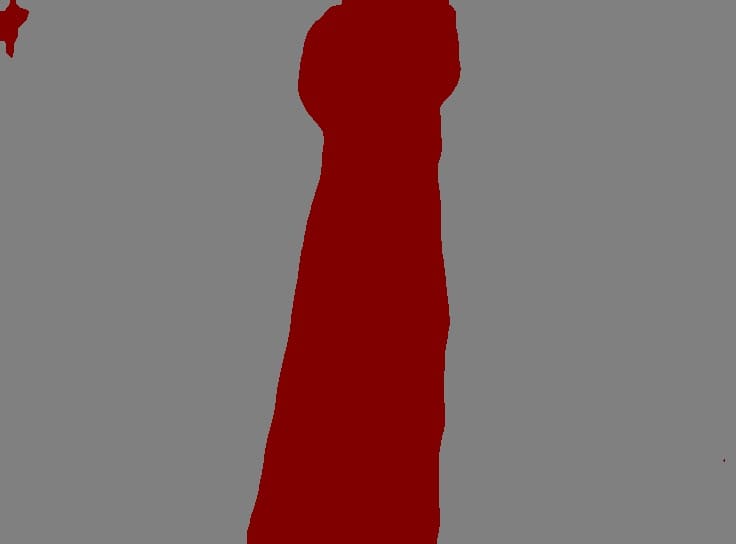}} \\
 \cmidrule(l{2em}r{2em}){1-4}
\multirow{3}{*}[0em]{Results}& Background [\%] &  91.76  & 91.21 \\ 
& Pipeline [\%]  & 66.42 &  60.12  \\ 
& mIoU [\%] & 79.09 &  75.66 \\ 
\bottomrule
\end{tabular}}
\end{table}

\subsection{RGB Image Segmentation}
 The models chosen for the RGB image segmentation experiments were SegFormer~\cite{xie2021segformer} and DeepLabV3~\cite{chen2017rethinking}. SegFormer is a visual transformer that uses transformer blocks in the encoder and uses the encoder's multiscale output features as inputs for a multilayer perceptron decoder. Using multiscale features in the decoder helps combine local and global information to achieve better results. This model was implemented in Pytorch and trained from scratch.\footnote{Reference implementation: \url{https://github.com/FrancescoSaverioZuppichini/SegFormer}} DeepLabV3 is a largely used \gls{CNN} for image segmentation that uses dilated convolution, which gives a better field of view for the filters, allowing for better detection of objects in different scales. This second model used the official PyTorch implementation with weights pre-trained on a subset of the COCO dataset. 

Both models were trained with cross-entropy loss, using the Adam optimizer with an initial learning rate of $10^{-4}$. The models were trained using the 647 annotated frames. The first 60\% labeled images from SubPipeMini were used for training, the next 20\% for validation, and the last 20\% for testing.  A suite of augmentation techniques was employed to mitigate overfitting, including flipping, rotation, cropping, and perturbations in RGB channels, saturation, and hue.

Even though SegFormer was trained from scratch, it still achieved good results, recognizing the pipeline relatively well. It indicates that the data has enough quantity and quality to train a deep-learning model. At the same time, even though the pipeline is a straight line, with an easy shape to be segmented, the \gls{IoU} for this class has room for improvement for both models, highlighting the difficulties of processing underwater images. Since pipeline tracking and inspection is a critical task, ideally, the \gls{IoU} for the pipeline should be improved. Perhaps pre-processing the images, using techniques to, e.g., increase the contrast, could improve the results. 

\subsection{Object Detection in Side-Scan Sonar images}
SubPipe provides SSS images featuring underwater pipelines. To evaluate the dataset's validity and quality, the YOLOX object detection model has been employed in two sizes: YOLOX-S with 9 million parameters and YOLOX-Nano with 0.9 million parameters~\cite{yoloxpaper}. The models were trained on high-frequency data from timestamp \texttt{1693573388} to \texttt{1693574657}, providing 15,404 images divided into 70\% for training, 15\% for testing, and 15\% for validation. Both models used pre-trained weights from the COCO dataset and trained during 300 epochs. The validation results on the training data are presented in the first column of \cref{table:object_detection}, indicating a 98\% accuracy for YOLOX-S and 96\% for YOLOX-Nano.

To further ensure the dataset's validity, a separate SSS mission from a different survey provided an additional validation set, from which 7,900 images were extracted. The second validation's results are in the second column of \cref{table:object_detection}.

\begin{figure}[!t]
\centering
{\begin{tabular}
{cc}
Training Data & Validation Data\\ 

\includegraphics[width=0.2\textwidth]{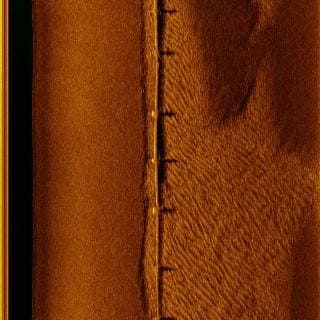} & \includegraphics[width=0.2\textwidth]{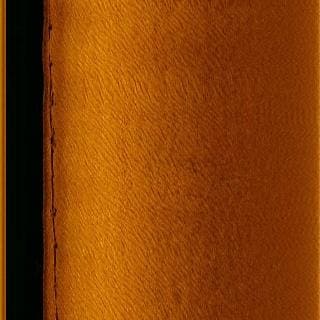}\\
\end{tabular}}
\caption{Comparison between training and validation datasets. The sample on the left illustrates the pipeline and its environment as represented in the training set, while the sample on the right depicts the validation dataset. The pipeline is next to the nadir gap in the second image.}
\label{fig:sample_SSS_dataset}
\end{figure}

A comparative analysis of both validation datasets revealed that neither model adequately detected objects in the second, validation dataset. Figure \ref{fig:sample_SSS_dataset} includes two samples from each dataset, illustrating that despite being collected from the same location, the datasets possess distinct characteristics, such as sand, pipeline shadow, and pipeline size variations. These characteristics implicitly offer insights into the models' learning patterns. Consequently, the experimental results suggest that even with a sufficient dataset size and high training metric values, environmental factors such as the sand's properties and the vehicle's altitude (affecting the pipeline shadow) must be considered for effective pipeline detection.

\begin{table}[!htbp]
\caption{Object Detection Results.}
\centering
\footnotesize
\label{table:object_detection}
\begin{tabular}{l c c}
\toprule
Model                                      & $\textit{AP}_{50-95}$ Training      & $\textit{AP}_{50-95}$ Validation\\    
\midrule
YOLOX-S            & 98.1\%    & 14.9\% \\
YOLOX-Nano  & 96.3\%    & 10.06\% \\
\bottomrule

\end{tabular}
\end{table}

\section{Conclusion and Future Work}
\label{sec:conclusion}

SubPipe is an underwater pipeline dataset for visual-inertial SLAM, object detection and segmentation tasks. 
It presents challenging imaging conditions characteristic of underwater environments. 
This paper proposes a set of experiments consisting of deploying state-of-the-art algorithms to benchmark those challenges. These experiments underscore the challenges and insights in deploying visual SLAM, image segmentation, and object detection in the context of submarine pipeline inspection. 

Geometry-based visual SLAM algorithms struggle with SubPipe's low-textured areas, leading to track loss or inadequate feature tracking, whereas the learning-based TartanVO shows promise in navigating these conditions. For RGB image segmentation, both SegFormer and DeepLabV3 show promising results despite the complexities of underwater imagery. 
Object detection experiments on SSS data demonstrate the influence of recording conditions on the performance of the YOLOX model. These experiments indicate that the model's accuracy may be compromised even when using the same SSS system, frequency, and pipeline, highlighting the inherent challenges in applying object detection to SSS images.

Collectively, these findings accentuate the promise of learning-based algorithms for underwater computer vision and the need for wider availability of data recorded under such conditions that allow more general models.
Consequently, the public release of this dataset serves as a critical contribution to addressing the scarcity of publicly available underwater datasets for computer vision.

\section*{Acknowledgments} 

This project has received funding from the European Union's Horizon 2020 research and innovation programme under the Marie Skłodowska-Curie grant agreement No. 956200. For further information, please visit \url{https://remaro.eu}.


\addtolength{\textheight}{-10.5cm}  







\bibliographystyle{ieeetr}
\bibliography{refs.bib}

\end{document}